\documentclass[sigconf]{acmart}

\usepackage{balance}

\def\BibTeX{{\rm B\kern-.05em{\sc i\kern-.025em b}\kern-.08emT\kern-.1667em\lower.7ex\hbox{E}\kern-.125emX}}

\usepackage{amsmath}
\DeclareMathOperator*{\argmax}{arg\,max}

\usepackage{booktabs} 
\usepackage{xcolor}
\usepackage[linesnumbered,ruled]{algorithm2e}
\usepackage{caption}
\usepackage{subcaption}
\usepackage{multirow}
\usepackage[noend]{algpseudocode}
\usepackage{mathtools}

\algnewcommand{\IfThenElse}[3]{
   \algorithmicif\ #1\ \algorithmicthen\ #2\ \algorithmicelse\ #3}
\newcommand{\model}{{ActSense}}
\renewcommand{\proofname}{Proof sketch}

\copyrightyear{2019}
\acmYear{2019}
\acmConference[CIKM '19]{The 28th ACM International Conference on Information and
Knowledge Management}{November 3--7, 2019}{Beijing, China}
\acmBooktitle{The 28th ACM International Conference on Information and Knowledge
Management (CIKM '19), November 3--7, 2019, Beijing, China}
\acmPrice{15.00}
\acmDOI{10.1145/3357384.3357929}
\acmISBN{978-1-4503-6976-3/19/11}

\begin{document}

\fancyhead{}

\title{Active Collaborative Sensing for Energy Breakdown}

\author{Yiling Jia}
\affiliation{%
\department{Department of Computer Science}
  \institution{University of Virginia}
  \city{Charlottesville}
  \state{VA}
  \country{USA}
}
\email{yj9xs@virginia.edu}

\author{Nipun Batra}
\affiliation{%
\department{Department of Computer Science}
  \institution{IIT Gandhinagar}
  \city{Gandhinagar}
  \country{India}
}
\email{nipun.batra@iitgn.ac.in}

\author{Hongning Wang}
\author{Kamin Whitehouse}
\affiliation{
\department{Department of Computer Science}
    \institution{University of Virginia} \city{Charlottesville}
  \state{VA}
  \country{USA}
 }
\email{{hw5x, whitehouse}@virginia.edu}



%
\renewcommand{\shortauthors}{Trovato and Tobin, et al.}

%
\begin{abstract}
Residential homes constitute roughly one-fourth of the total energy usage worldwide. Providing appliance-level energy breakdown has been shown to induce positive behavioral changes that can reduce energy consumption by 15\%. Existing approaches for energy breakdown either require hardware installation in every target home or demand a large set of energy sensor data available for model training. However, very few homes in the world have installed sub-meters (sensors measuring individual appliance energy); and the cost of retrofitting a home with extensive sub-metering eats into the funds available for energy saving retrofits. As a result, strategically deploying sensing hardware to maximize the reconstruction accuracy of sub-metered readings in non-instrumented homes while minimizing deployment costs becomes necessary and promising.

In this work, we develop an active learning solution based on low-rank tensor completion for energy breakdown. We propose to actively deploy energy sensors to appliances from selected homes, with a goal to improve the prediction accuracy of the completed tensor with minimum sensor deployment cost. 
We empirically evaluate our approach on the largest public energy dataset collected in Austin, Texas, USA, from 2013 to 2017. The results show that our approach gives better performance with fixed number of sensors installed, when compared to the state-of-the-art, which is also proven by our theoretical analysis.

\end{abstract}

%
%

\ccsdesc[500]{Computing methodologies~Active learning settings}
\ccsdesc[300]{Human-centered computing~Ubiquitous computing}
%
\keywords{active learning, tensor completion, energy breakdown}

%
\maketitle

\section{Introduction}
\label{sec:intro}
Residential homes are one of the largest energy consumers worldwide, constituting roughly one-fourth of the total
energy usage \cite{perez2008review}. Part of this energy could be saved by providing an energy breakdown, i.e., per-appliance energy consumption summary. Studies have shown that energy breakdown enables informed decision making by different actors in the home's energy ecosystem \cite{armel_2013}. For example, studies~\cite{kelly2014nilmtk, armel_2013} report energy feedback causes behavioral changes that can reduce energy consumption by 15\%. It also helps power utility companies and policymakers to improve load forecasting \cite{armel_2013}, detect broken or mis-configured equipment \cite{katipamula2005review}, and target the most inefficient homes for energy efficiency programs.

Various energy breakdown techniques have been proposed in the past, such as direct sensing systems~\cite{jiang2009design, debruin2015powerblade} and non-intrusive load monitoring (NILM)~\cite{hart_1992, fhmm, kolter_2010, kolter_2012}. Different from those models which require hardware to be installed in every home, recently, collaborative sensing~\cite{batra2015good, batra2017systems, batra2018transferring} has attracted increasing attention due to its low cost and high scalability. Collaborative sensing, which aims at reconstructing the appliance-level energy data of non-instrumented homes based on data collected from other homes, only requires the easily available information of non-instrumented homes, such as the monthly energy bills, square footage, and number of occupants. The basic premise is that, while every home is unique, the common design and construction patterns among homes create a shared and repeating structure, which gives rise to a sparse set of factors contributing to energy variation across homes. A typical approach is to factorize energy readings into a low-dimensional space, and predict energy consumption in a non-instrumented home with this low-dimensional model based on the high-fidelity data collected in other homes. It is worth mentioning that collaborative sensing algorithms only perform at a low temporal resolution, such as monthly data, unlike NILM-type solutions that can produce a high frequency appliance energy time-series. However, previous studies have shown sustained savings even at a monthly resolution~\cite{faustine2017survey, kelly2016does}, which support the value of low frequency energy breakdown.

While collaborative sensing alleviates the scalability issue of NILM-type methods by removing the requirement of per-home instrumentation, these solutions have assumed the existence of relevant training data, i.e., appliance-level energy readings, from some fully instrumented homes. But, in reality, very few homes in the world have been instrumented with sub-meters (appliance level energy meters). As a result, sensor deployment is still inevitable to apply such methods. In this work, to further improve the scalability of collaborative sensing, we seek to answer the question: \emph{can we minimize the deployment cost by selectively deploying sensing hardware to a subset of homes and appliances while maximizing the reconstruction accuracy of sub-metered readings in non-instrumented homes?} We name this new research problem as active sensor deployment for energy breakdown.

Active sensor deployment for energy breakdown differs from classical active learning problems \cite{settles2012active, tong2001support, cohn1996active} in three major aspects. First, energy readings are time-series data. New readings are constantly generated, and they are influenced by various external and internal factors, such as season \cite{batra2018transferring} and occupant activities. The typically imposed assumption in active learning literature that observations are independent and identically distributed no longer holds in this situation. 
Second, once a sensor has been installed in a home, the monitored appliance readings from that home will become available thereafter. This directly introduces the explore/exploit dilemma in active sensor deployment, because one has to balance the choice of instrumentation that focuses on the current bottleneck of reconstruction accuracy, and that improves model accuracy for future predictions. For example, in spring, furnace might consume most energy per-household for heating, and therefore more instrumentation on furnace is required to obtain a more accurate model. But in summer, air conditioning system will become the main source of energy consumption. If one has not instrumented any air conditioning system before, the reconstruction accuracy on it will be poor. It is necessary to plan the instrumentation ahead of time, so as to obtain a high accuracy model before the consumption peaks.
Third, an instrumentation choice concerns two different types of entities, i.e., homes and appliances, which are not independent. This adds another dimension into the explore/exploit dilemma for sequential decision making: which <home, appliance> pair to instrument next for maximizing future reconstruction accuracy. 

 \begin{figure}[t]
    \centering
    \includegraphics[width=0.8\linewidth]{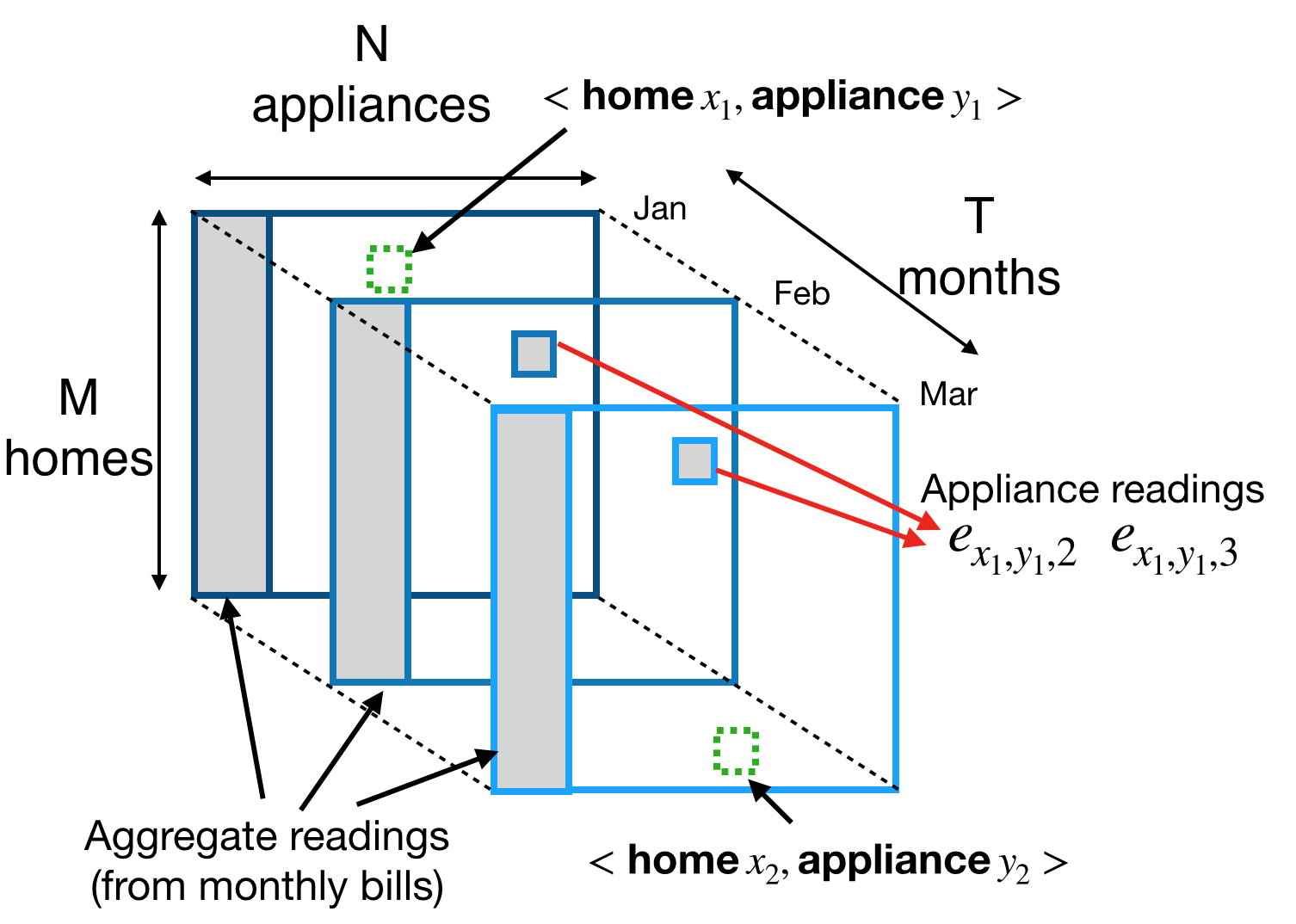}
    \vspace{-2mm}
    \caption{Active sensor deployment for energy breakdown. (1) We treat aggregate readings of each home as a special appliance, and they are always available; (2) At the end of January, if <home $x_1$, appliance $y_1$> is selected, appliance readings $\mathbf{e}_{x_1, y_1, t}$ become available for $t \geq 2$. 
    }
    \label{fig:problem}
\end{figure}

In this work, we follow factorization based collaborative sensing \cite{batra2018transferring, batra2017systems}, and propose to perform active sensor deployment via active tensor completion. 
First of all, we view energy readings as a three-way tensor as illustrated in Figure \ref{fig:problem}: with homes, appliances and time as three separate dimensions. In this tensor, a cell is filled with an observation if the corresponding appliance has been monitored in this home before; and predictions are made in cells with missing values. Although the size of this tensor is large and increasing, due to the continuously observed energy readings, we believe the tensor is low rank. Learning a low rank representation for homes and appliances enables us to predict energy use in those non-instrumented homes. Evidence of this modeling assumption has been gathered in various existing studies \cite{batra2017matrix,batra2018transferring}. 
As a result, the active sensor deployment problem can be naturally formalized as an active tensor completion problem: deciding which cells in the tensor to query so that the reconstruction accuracy can be maximized. Specifically, at the end of each month, we query the home and appliance pairs that have the highest uncertainty in the current tensor reconstruction, which we prove to reduce reconstruction uncertainty mostly rapidly. And to project a model's prediction uncertainty of future readings in a longer term, we incorporate external seasonal information into model estimation. This helps the model react to future season changes earlier. We name our solution as Active Collaborative Sensing, or \model{} in short. 

We rigorously prove that with high probability the developed algorithm achieves a considerable reduction in accumulated estimation error. In other words, it requires less sensor deployment to obtain the same level of reconstruction accuracy compared to any other sensor installation. We evaluate \model{} on a publicly available dataset called Dataport~\cite{parson2015dataport}, which is the largest public energy dataset collected in U.S. With a fixed budget of sensors, our model achieves better energy breakdown performance compared to three baseline approaches. Besides, extensive analysis of the experiments shows that integrating the temporal seasonal information can help to foresee the energy usage trends and prepare the sensor installation in advance.
\section{Related Work}
\label{sec:relwork}
Since George Hart's seminal work on non-intrusive load monitoring (NILM) in the early 1990s~\cite{hart_1992}, the research community has proposed several solutions to scale up energy breakdown. 
The basic idea of NILM algorithms is to perform source separation on the power signal measured at a single point (home mains)~\cite{zoha, armel_2013}. Majority of these algorithms worked on time-series data obtained from a smart meter, collected at rates from 10s of kHz to a reading once every hour~\cite{parson2012non, kolter_2010, kolter_2012, shao2013temporal}. But NILM type algorithms still require instrumentation across each home.
Recently, there is a line of research collectively referred to as collaborative sensing that focuses on energy breakdown without any hardware installation in a test home~\cite{batra2015good, batra2017matrix, batra2018transferring}. Their key idea is that ``similar homes would have a similar per-appliance energy consumption''. These approaches 
estimate the energy breakdown of a home by finding a similar home (based on monthly energy use) that already has an energy breakdown available. 
These approaches were shown to be scalable and accurate compared to the state-of-the-art NILM approaches. However, such algorithms require available energy breakdown data for model learning; they cannot apply when this data is missing. 


The active sensor deployment problem has been studied in robotics for several decades. 
But the main focus there is on active sensing in a fixed region by utilizing the spatial information.
For example, in \cite{chen2011active}, the vision sensor was purposefully configured and placed at several positions to observe a target.
Ksenia et al. ~\cite{shubina2010visual} addressed the object search problem 
in surveillance by optimizing the probability of finding the target given a fixed cost limit. 
In \cite{banta2000next,chen2005vision, pito1999solution}, the authors proposed the approaches with next best view planning methods to improve object modeling and recognition, while several studies have been explored on active sensing based on information entropy and some rule-based methods~\cite{li2005information, kutulakos1995global}. 
But in active energy breakdown problem that we study in this paper, the energy data is a time series generated continuously from different homes and appliances. Temporal pattern in energy consumption across different homes and appliances is the key for optimizing sensor deployment. To the best of our knowledge, no existing work addresses the active sensing problem in energy breakdown.

On a related direction of research, active matrix completion has gained increasing attention in the past decade, due to the wide adoption of matrix completion based collaborative filtering algorithms. The essence of active matrix completion is to assess statistical uncertainty of each unobserved cell in the matrix, and propose to query the most informative ones to improve the factorization model's prediction accuracy. Variational Bayesian Inference (VBI)~\cite{beal2003variational} is one of the most popularly used methods for uncertainty assessment. There exist several solutions leveraging VBI to estimate model uncertainty and using mutual information~\cite{silva2012active}, prediction variance~\cite{sutherland2013active} and information gain~\cite{houlsby2014cold} to query the observations actively. Shayok et al. utilized graphical lasso for model estimation and queried instances with the highest uncertainty\cite{chakraborty2013active}. Some recent developments solve online matrix completion with multi-armed bandit algorithms, a reference solution for explore-exploit trade-off~\cite{auer1995gambling, auer2002finite, li2010contextual}. Model uncertainty is assessed by posterior sampling \cite{kawale2015efficient} or estimation of prediction's confidence interval \cite{wang2017factorization}.
However, in the problem of energy breakdown, it is insufficient to view the energy data as a matrix, as the data is continuously generated over time. Observations collected from different time periods now become depend on each other. An optimal query strategy should take this temporal dependency into consideration to balance the uncertainty from different homes and appliances over time. 
This, to the best of our knowledge, has not been carefully studied to date.
\section{Methodology}
\label{sec:method}
There are two essential research questions in active sensor deployment for energy breakdown: (1) how to accurately perform energy breakdown on a given set of homes with sub-metering data; and (2) how to select <home, appliance> pairs for additional instrumentation to improve future energy breakdown accuracy. In this section, we first provide a detailed problem definition, then elaborate the design of our proposed Active Collaborative Sensing solution, and provide theoretical analysis of its convergency in the end.

\subsection{Problem Statement}
\label{sec:prob}
In active sensor deployment, we aim to maximize the energy breakdown accuracy while minimizing the sensor deployment cost. We formally define our energy reading data as a three-way tensor, $\mathbf{E_{M \times N \times T}}$, where the elements contain energy readings from $\mathbf{M}$ homes and $\mathbf{N}$ appliances for up to  $\mathbf{T}$ months. Detailed reading is available in a cell when the corresponding appliance is monitored in that home beforehand; otherwise the reading is unknown. 
There are some special structural properties of this tensor, which make the active sensor deployment problem especially challenging.

\noindent{\bf $\bullet~$ Time-series data.} In energy breakdown scenario, energy data is continuously generated and collected after every sampling cycle, e.g., monthly in Figure \ref{fig:problem}. 
At the end of each month, new <home, appliance> pairs can be 
selected according to predefined selection strategy and sensor deployment budget.

\noindent{\bf $\bullet~$ Sensor installation.}
In sensor deployment, once a sensor is installed, the appliance readings will always be available in the future. For example, if we install $K$ sensors in January and $K$ sensors in February, we will have $K$ extra readings at the end of January and $2 \times K$ extra readings at the end of February. 

\noindent{\bf $\bullet~$ Aggregate readings.}
We view the household aggregate energy as a ``special'' appliance in the energy tensor, which is always available in the form of monthly energy bill.


The active sensor deployment for energy breakdown can be formalized as an active tensor completion problem. More specifically, the learning procedure is: at month $t \in [1, \mathbf{T}-1]$, based on the observed energy tensor $\mathbf{E_{\Omega} \in R^{M \times N \times t}}$ where $\mathbf{\Omega}$ represents the set of indices of observed elements, and subject to budget constraint, 
we want to install sensors to a set of <home, appliance> pairs which will bring the largest reconstruction error reduction of the energy tensor in the future months.

\subsection{Energy Tensor Completion}
\label{sec:tensor}
The basic factorization model we adopt here is the Canonical Polyadic (CP) decomposition, which is also known as rank decomposition. Our core intuition is that the common design and construction patterns in residential homes create a shared and repeating structure. And the heterogeneity across homes can be captured by a low dimension representation of individual homes \cite{batra2018transferring}. Examples of such low-dimensional representations could be home insulation or the number of occupants. Similarly, the energy consumption dependence of different appliances with respect to home design and weather pattern can also be encoded using a low-dimensional representations. 

Therefore, we can decompose the energy tensor $\mathbf{E_{M \times N \times T}}$ into three factors: home factor $\mathbf{H} \in \mathbb{R}_+^{M \times r}$, appliance factor $\mathbf{A} \in \mathbb{R}_+^{N \times r}$ and season factor $\mathbf{S} \in \mathbb{R}_+^{T \times r}$, where $\mathbf{H}$, $\mathbf{A}$ and $\mathbf{S}$ are independent non-negative matrices and $r$ is the rank of the energy tensor. 
With such a decomposition design, the $i$th row of $\mathbf{H}$, denoted as $\mathbf{h}_i$, is the latent factor of the $i$th home; the $j$th row of $\mathbf{A}$, denoted as $\mathbf{a}_j$, is the latent factor for $j$th appliance; and the $k$th row of $\mathbf{S}$, denoted as $\mathbf{s}_k$, is the latent factor for $k$th month.
To reflect the fact that energy readings should be finite in all homes and appliances, we further assume that the L2 norm of these latent matrices are upper bounded: $\forall i, ||\mathbf{h}_i||_2 \leq P$; $\forall j, ||\mathbf{a}_j||_2 \leq Q$; and $\forall k, ||\mathbf{s}_k||_2 \leq R$.
The energy consumption of appliance $j$ in home $i$ at month $k$ can be estimated by $\mathbb{E}[\mathbf{e}_{ijk}]=<\mathbf{h}_i, \mathbf{a}_j, \mathbf{s}_k>$, where $\langle\cdot{,}\cdot{,}\cdot\rangle$ represents the triple product. 
In each month $t$, in order to estimate the latent factors, regularized quadratic loss over the observations in the energy tensor $\mathbf{E_{M \times N \times t}}$ is employed. The objective function for latent factor estimation at month $t$ is,
\begin{align}
\label{eqn:obj}
    F =& \sum_{(i, j, k) \in \Omega_t}(<\mathbf{h}_{i}^t, \mathbf{a}_{j}^t, \mathbf{s}_{k}^t> - \mathbf{e}_{ijk})^2 \\ 
    &+ \lambda_1\sum_{i=1}^M||\mathbf{h}_{i}^t||_2 + \lambda_2\sum_{j=1}^N||\mathbf{a}_{j}^t||_2 + \lambda_3\sum_{k=1}^t||\mathbf{s}_{k}^t||_2 
    \nonumber
\end{align}
where $\Omega_t$ represents the set of indices of observed energy readings till month $t$, and $\lambda_1$, $\lambda_2$ and $\lambda_3$ are the coefficients for L2 regularization. The notion of $(\mathbf{h}_{i}^t, \mathbf{a}_{j}^t, \mathbf{s}_{k}^t)$ denotes the latest estimate of latent factors at month $t$.
The inclusion of L2 regularization terms is critical to our solution in two folds. First, it makes the sub-problems in coordinate decent based optimization well-posed, so that we have closed form solutions for the latent factors at each round. Second, it helps to remove the scaling indeterminacy among the estimates of those latent factors.

We estimate the latent factors with alternative least square (ALS).
Specifically, the closed-form estimation of $(\mathbf{h}_i, \mathbf{a}_j, \mathbf{s}_k)$ with respect to Eq~\eqref{eqn:obj} at month $t$ can be achieved by $\hat{\mathbf{h}}_{i}^t = {(\mathbf{A}_{i}^t})^{-1}\mathbf{b}_{i}^t$, ${\hat{\mathbf{a}}_{j}^t} = {(\mathbf{C}_{j}^t})^{-1}\mathbf{d}_{j}^t$ and $\hat{\mathbf{s}}_k^t = {(\mathbf{E}_{k}^t})^{-1}\mathbf{f}_{k}^t$, in which,
\begin{align}
    \mathbf{A}_{i}^t =& \lambda_1\mathbb{I}_1 + \sum_{\mathclap{(i, j, k)\in \Omega_t}} (\mathbf{a}_{j}^t \circ \mathbf{s}_k^t)(\mathbf{a}_{j}^t \circ \mathbf{s}_k^t)^\top {,}\; \mathbf{b}_{i}^t = \sum_{\mathclap{(i, j, k) \in \Omega_t}} \mathbf{e}_{ijk}(\mathbf{a}_j^t \circ \mathbf{s}_k^t) \label{eqn:Ab} \\
    \mathbf{C}_{j}^t =& \lambda_2\mathbb{I}_2 + \sum_{\mathclap{(i, j, k)\in \Omega_t}} (\mathbf{h}_i^t \circ \mathbf{s}_k^t)(\mathbf{h}_i^t \circ \mathbf{s}_k^t)^\top {,}\; \mathbf{d}_{j}^t = \sum_{\mathclap{(i, j, k) \in \Omega_t}} \mathbf{e}_{ijk}(\mathbf{h}_i^t \circ \mathbf{s}_k^t) \label{eqn:Cd}\\
    \mathbf{E}_{k}^t =& \lambda_3\mathbb{I}_3 + \sum_{\mathclap{(i, j, k)\in \Omega_t}} (\mathbf{h}_i^t \circ \mathbf{a}_j^t)(\mathbf{h}_i^t \circ \mathbf{a}_j^t)^\top {,}\; \mathbf{f}_{k}^t = \sum_{\mathclap{(i, j, k) \in \Omega_t}} \mathbf{e}_{ijk}(\mathbf{h}_i^t \circ \mathbf{a}_j^t) \label{eqn:Ef}
\end{align}
where $\mathbb{I}_1$, $\mathbb{I}_2$ and $\mathbb{I}_3$ are identity matrices with dimensions of $r \times r$. The operation $ (\circ) $ represents the element-wise product of two vectors, and it is easy to verify that $<\mathbf{h}_i, \mathbf{a}_j, \mathbf{s}_k> = (\mathbf{h}_i \circ \mathbf{a}_j)^T\mathbf{s}_k = (\mathbf{h}_i\circ \mathbf{s}_k)^T\mathbf{a}_j = (\mathbf{a}_j \circ \mathbf{s}_k)^T\mathbf{h}_i$. Proper projection is needed to ensure the estimated factors are non-negative and their norms are in the required range. The estimated factors predict the unseen values in the energy tensor at month $t$ by $\hat{\mathbf{e}}_{ijk}= <\hat{\mathbf{h}}_i^t, \hat{\mathbf{a}}^t_j, \hat{\mathbf{s}}^t_k>$. 


\subsection{Uncertainty Quantification}
\label{sec:uncertainty}




With the learnt latent factors, household energy breakdown can be readily provided by the aforementioned factorization-based solution. 
However, without sufficient appliance-level energy readings, the estimated latent factors are subject to various sources of uncertainty, e.g., variance in energy use or errors in sensor readings, which directly lead to volatility in appliance-level energy prediction. Thus, to improve the estimation quality of future energy consumption and obtain strong performance guarantee, strategically querying observations from the non-instrumented home and appliance pairs is of paramount importance. In \model{}, we propose to select the unobserved pairs with the highest estimation uncertainty so as to bring the largest reconstruction error reduction.

The first step in active selection is to quantify the estimation uncertainty in the energy tensor completion. The main reason for the estimation uncertainty is the existence of potential noise in the observed energy readings, e.g., imperfect sensor hardware, unexpected energy use (such as energy consumed in wire transition), and etc. Passing through the factorization procedure, such noise leads to uncertainty in latent factor estimation, e.g., $||\hat{\mathbf{h}}_i^t - \mathbf{h}_{i}^*|| \neq 0$, where $\mathbf{h}_{i}^*$ is the ground-truth latent home factor, and $\hat{\mathbf{h}}_i^t$ is the home factor estimated with the observed noisy energy readings $\mathbf{e}^{obs}$ at month $t$. The discrepancy between the estimated and ground-truth latent factors directly contributes to the uncertainty in appliance-level energy consumption estimation. 
To model uncertainty, we assume the noise follows a zero-mean Gaussian distribution: $\mathbf{e}_{ijk}^{obs} = \mathbf{e}_{ijk}^* + \eta_{ijk} = <\mathbf{h}_i^*, \mathbf{a}_j^*, \mathbf{s}_k^*> + \eta_{ijk}$, where $\eta_{ijk} \sim \mathcal{N}(0,\,\sigma^{2}_{ijk})$ is the noise term, $e_{ijk}^{obs}$ is the observed energy reading, $e_{ijk}^*$ is the noise-free energy consumption, and $h_i^*$, $a_j^*$, and $s_k^*$ are the ground-truth latent factors. Under the context of tensor completion for energy breakdown, at month $t$, the uncertainty of appliance-level energy estimation comes from the estimation error of latent home factors, i.e., $||\hat{\mathbf{h}}_i^t - \mathbf{h}_{i}^*||$,  appliance factors, i.e., $||\hat{\mathbf{a}}_j^t - \mathbf{a}_j^*||$, and season factors, i.e., $||\hat{\mathbf{s}}_k^t - \mathbf{s}_k^*||$, caused by the noise. With the closed-form solution in our coordinate descent estimation, at each time $t$, the confidence interval of estimated $\hat{\mathbf{h}}_i^t$, $\hat{\mathbf{a}}_j^t$, and $\hat{\mathbf{s}}_k^t$ can be analytically computed by the following lemma.

\begin{lemma}
With proper initialization of the coordinate descent estimation, the Hessian matrix is positive definite at the optimizer $\mathbf{h}^*_i$, $\mathbf{a}^*_j$, and $\mathbf{s}^*_k$. Thus, for any $\epsilon_1 > 0$. $\epsilon_2 > 0$, $\epsilon_3 > 0$, and $\delta \in (0, 1)$, with probability at least $1 - \delta$, the estimated error of latent factors satisfies,

\begin{align*}
    ||\hat{\mathbf{h}}_i^t - \mathbf{h}^*_i||_{\mathbf{A}_i^t} 
    &\leq \sqrt{r\ln{\frac{\lambda_1r+|\Omega_t|Q^2R^2}{\lambda_1 \cdot r \cdot \delta}}} + \sqrt{\lambda_1}P + \frac{2PQ^2R^2}{\sqrt{\lambda_1}}(G_2 + G_3) \\
    ||\hat{\mathbf{a}}_j^t - \mathbf{a}^*_j||_{\mathbf{C}_j^t} 
    &\leq \sqrt{r\ln{\frac{\lambda_2r+|\Omega_t|P^2R^2}{\lambda_2 \cdot r \cdot \delta}}} + \sqrt{\lambda_2}Q + \frac{2P^2QR^2}{\sqrt{\lambda_1}}(G_1 + G_3) \\
    ||\hat{\mathbf{s}}_k^t - \mathbf{s}^*_k||_{\mathbf{E}_k^t} 
    &\leq \sqrt{r\ln{\frac{\lambda_3r+|\Omega_t|P^2Q^2}{\lambda_3 \cdot r \cdot \delta}}} + \sqrt{\lambda_3}R + \frac{2P^2Q^2R}{\sqrt{\lambda_3}}(G_1 + G_2) 
\end{align*}
where,
\begin{alignat*}{3}
    G_1&= \frac{f_1(1 - f_1^{|\Omega_t|})}{1 - f_1} \,{,}\, &&G_2 = \frac{f_2(1 - f_2^{|\Omega_t|})}{1 - f_2} \,{,}\, & &G_3 = \frac{f_3(1 - f_3^{|\Omega_t|})}{1 - f_3} \\
    f_1&= q_1 + \epsilon_1 \,{,}\, &&f2 = q_2 + \epsilon_2 \,{,}\, & &f3 = q_3 + \epsilon_3
\end{alignat*}
and $q_1 \in (0, 1)$, $q_2 \in (0, 1)$, $q_3 \in (0, 1)$, $|\Omega|$ is the cardinality of set $\Omega$.
\label{lemma}
\end{lemma}

The detailed proof of Lemma~\ref{lemma} can be found in the provided supplementary material. With Lemma~\ref{lemma}, we obtain a tight construction for the estimation uncertainty of the latent factors, which can be easily transformed to the uncertainty of energy tensor estimation $\hat{\mathbf{e}}_{ijk}$. 
As described in Section~\ref{sec:prob}, in active energy tensor completion, the <home, appliance> pairs are selected at the end of each month. Under such a setting, we do not have a choice of when to install the sensors. Therefore, we define $\alpha_x^t$ and $\alpha_y^t$ as the upper bound of $||\hat{\mathbf{h}}_i^t - \mathbf{h}^*_i||_{\mathbf{A}_i^t}$ and $||\hat{\mathbf{a}}_j^t - \mathbf{a}^*_j||_{\mathbf{C}_j^t}$, and propose the following <home, appliance> pair selection strategy at month $t$:
\begin{align}
\label{eqn:select}
    (x, y) = \argmax_{x \in [1, M], y \in [1, N]}
   \bigg(&\alpha_x^t\sqrt{({\mathbf{a}_y^t}\circ \hat{\mathbf{s}}_{t}^t)^\top{\mathbf{A}_x^t}^{-1}({\mathbf{a}_y^t}\circ \hat{\mathbf{s}}_{t}^t)} \\
   +& \alpha_y^t\sqrt{({\mathbf{h}_x^t}\circ \hat{\mathbf{s}}_{t}^t)^\top{\mathbf{C}_y^t}^{-1}({\mathbf{h}_x^t}\circ \hat{\mathbf{s}}_{t}^t)}\bigg)
   \nonumber
\end{align}
The detailed derivation of this uncertainty estimation is provided in the supplementary material. 
In Eq (\ref{eqn:select}), the two terms on the right-hand side are related to the estimation uncertainty of latent factor $\hat{\mathbf{h}}_{i}$ and $\hat{\mathbf{a}}_j$ at current month $t$. We choose to select the pairs with the highest estimation uncertainty of the associated latent factors, which leads to the best reduction in the model's overall prediction uncertainty. We prove this conclusion in Section \ref{sec:converge}.


\subsection{Active Collaborative Sensing}
\label{sec_actsense}
The aforementioned uncertainty estimation is constructed by the estimated season factor $\mathbf{s}^t_t$ of month $t$; hence, it only measures the model's prediction uncertainty at that particular month. But it does not consider the specific challenges we face in the context of energy breakdown (as discussed in Section \ref{sec:prob}). 
Basically, there exists a time lag in sensor deployment: all the decisions we make are based on the current knowledge of energy consumption, but the target of instrumentation is to get a better energy breakdown accuracy for future predictions. We can only verify our decisions at least one month later. The active selection will be more effective if we could foresee the changes in future and prepare the sensor installation accordingly. However, with limited observations, it is hard to extrapolate the usage pattern of different appliances across homes nor to make a reasonable prediction of future use, not to mention the uncertainty estimation of this future prediction. 

Motivated by the analysis in previous literature~\cite{batra2018transferring}, within one geo-region, the season factors learned from energy data are highly correlated with the region's seasonal pattern.
Previous work also showed that the season patterns are similar and repeating across years. Hence, with the aggregate readings collected from the monthly bills of the past year, we can obtain a rough estimation of the season factors of the current year.
We then inject historically learned season factors into our latent factor estimation by changing Eq (\ref{eqn:obj}) to,
\begin{align}
\label{eqn:obj_new}
    F &= \sum_{(i, j, k) \in \Omega_t}(<\mathbf{h}_i^t, \mathbf{a}_j^t, \mathbf{s}_k^t> - \mathbf{e}_{ijk})^2 + \lambda_1\sum_{i=1}^M||\mathbf{h}_i^t||_2 \\
    &+ \lambda_2\sum_{j=1}^N||\mathbf{a}_j^t||_2 + \lambda_3\sum_{k=1}^t||\mathbf{s}_k^t - \mathbf{s}_{k}^{prev}||_2 
    \nonumber
\end{align}
\normalsize
where $\mathbf{s}^{prev}$ is the season factor learned from past years. As $\mathbf{s}^{prev}$ can be estimated from aggregate readings alone, we even do not require any instrumented homes from past years for this estimation. 
This new regularization term assumes that the season across years is similar, and it changes smoothly between two adjacent years. As a result, when the observations from future months are not yet available, we can directly use the previously estimated season factors as an estimate for those future months. With these estimated season factors, we can extend Eq \eqref{eqn:select} to quantify the model's prediction uncertainty for future months. 

On the other hand, because at the end of each month we update the latent factor estimations based on the latest available energy readings, we would have an updated uncertainty estimation on the <home, appliance> pairs in the past months as well (by using the updated season factors for those months in Eq \eqref{eqn:select}), if we have not instrumented them yet. It provides us information about where our model is still uncertain in historical observations. It is therefore necessary to combine the uncertainty estimations about past, current, and future energy use 
to guide active sensor deployment. We use a time decay kernel to integrate those uncertainty estimations, and describe the procedure in Algorithm~\ref{algo:algo2}.

\begin{algorithm}[t]
\caption{Uncertainty(x, y, t)}
\label{algo:algo2}
    \SetKwInOut{Input}{Input}
    \SetKwInOut{Output}{Output}
    
    \Input{ $\mathbf{h}_x^t$, $\mathbf{a}_y^t$, $\mathbf{s}_{1:t}^t$, $\mathbf{A}_x^t$, $\mathbf{C}_y^t$,  $\mathbf{S}^{prev}$, $\alpha_x$, $\alpha_y$}
    \Output{$U(x, y, t)$}
    $U(x, y, t) = 0$ \\
    \For{$t'=1$ \KwTo $T$}{
        \IfThenElse{$t' \leq t$}{$\Tilde{s} \leftarrow \mathbf{s}_{t'}^t$}{$\Tilde{s} \leftarrow \mathbf{s}^{prev}_{t'}$}

        $U(x, y, t) \leftarrow U(x, y, t) + \textbf{Weight}(t', t) \cdot \Big(\alpha_x^t\sqrt{({\mathbf{a}_y^t}\circ \Tilde{\mathbf{s}})^\top(\mathbf{A}_x^t)^{-1}({\mathbf{a}_y^t}\circ \Tilde{\mathbf{s}})} + \alpha_y^t\sqrt{({\mathbf{h}_x^t}\circ \Tilde{\mathbf{s}})^\top(\mathbf{C}_y^t)^{-1}({\mathbf{h}_x^t}\circ \Tilde{\mathbf{s}})}\Big)$ 
    }
\end{algorithm}

As shown in Algorithm~\ref{algo:algo2}, the final uncertainty of a <home, appliance> pair is a linear combination of uncertainties across a fixed period of time, i.e., 12 months. 
The \textbf{Weight}$(t,t')$ function controls how much a particular month's uncertainty estimation contributes to the final decision. We use a triangle kernel to calculate the weight:
\begin{equation}
\label{eq_weighting}
    k(t', t) = \left\{\begin{matrix}
1 - \frac{|t' - t|}{\sigma} & if |t' - t| \leq \sigma\\ 
 0& otherwise 
\end{matrix}\right.
\end{equation}
where $t$ is the index of current month, and $t'$ is the index of a candidate month, and $\sigma$ is the parameter to control time decay. We always give the current month itself the highest weight. The intuition behind Eq \eqref{eq_weighting} is that as the season is expected to smoothly change between adjacent months, the estimation uncertainty from nearby months should be more important for the decision of sensor deployment. For example, in May, the uncertainty estimated from February should be less important than that from June.

Combining the integrated uncertainty estimation and the estimation of latent factors discussed in Section \ref{sec:tensor}, we can effectively perform active tensor completion, and therefore address the problem of active sensor deployment for energy breakdown. We name the resulting algorithm Active Collaborative Sensing, or \model{} in short, and illustrate the detailed procedure of it in Algorithm~\ref{algo:algo1}. Here, we assume uniform cost for instrumenting different appliances. Thus, considering a practical setting in active sensor deployment, at the end of each month, $L$ <home, appliance> pairs with the highest uncertainties will be selected for sensor installation, where $L$ can be determined by the total cost of instrumentation and budget. And we leave the work of measuring the various cost of sensor installation across different appliances as our future work.


    


\begin{algorithm}[t]
\caption{Active Collaborative Sensing}
\label{algo:algo1}

    \SetKwInOut{Input}{Input}
    \SetKwInOut{Output}{Output}

    \Input{$\lambda_1$, $\lambda_2$, $\lambda_3$ $\in$ $(0, \infty), L$}
    
    \lForEach{$i=1$ \KwTo $\mathbf{M}$}{
        $\mathbf{A}_{i}^0 \leftarrow \lambda_1 \mathbf{I}, \mathbf{b}_{i}^0 \leftarrow \mathbf{0}, \mathbf{h}_{i}^0 \leftarrow \mathbf{0}$
    }
    \lForEach{$j=1$ \KwTo $N$}{
        $\mathbf{C}_{j}^0\leftarrow \lambda_2 \mathbf{I}, \mathbf{d}_{j}^0 \leftarrow \mathbf{0}, \mathbf{a}_{j}^0 \leftarrow \mathbf{0}$
    }
    \lForEach{$t=1$ \KwTo $T$}{$\mathbf{E}_{t}^0 \leftarrow \lambda_2 \mathbf{I}, \mathbf{f}_{t}^0 \leftarrow \mathbf{0}, \mathbf{s}_{t}^0 \leftarrow \mathbf{0}$}
     
    Initialize the observation set $\Theta_0 \leftarrow \emptyset$ \;
    \For{$t=1$ \KwTo $T$}{
        Update $\Theta_t$ with monthly bills and sub-meter readings. \;
        \For{$\mathbf{e}_{ijk}$ from home $i$, appliance $j$ at time $k$ where $\mathbf{e}_{ijk} \in \Theta_t$}{
            Update $\mathbf{A}_i^t$, $\mathbf{b}_i^t$ by Eq~(\ref{eqn:Ab}), $\mathbf{h}_i^t \leftarrow (\mathbf{A}_i^t)^{-1}\mathbf{b}_i^t$ \;
            Update $\mathbf{C}_j^t$, $\mathbf{d}_j^t$ by Eq~(\ref{eqn:Cd}), $\mathbf{a}_j^t \leftarrow (\mathbf{C}_j^t)^{-1}\mathbf{d}_j^t$ \;
            Update $\mathbf{E}_k^t$, $\mathbf{f}_k^t$ by Eq~(\ref{eqn:Ef}), $\mathbf{s}_k^t \leftarrow (\mathbf{E}_k^t)^{-1}\mathbf{f}_k^t$\;
        }
        Install sub-meters to $L$ <home, appliance> pairs with highest \textbf{Uncertainty}(x, y, t)\;
    }
\end{algorithm}

\subsection{Convergence Analysis}
\label{sec:converge}
In this section, we theoretically analyze the sample selection strategy in our proposed Active Collaborative Sensing algorithm.

Based on Lemma~\ref{lemma} and the $q$-linear convergence property of alternative least square~\cite{uschmajew2012local}, at month $t$, for any home $i$, appliance $j$, and any month $k$ before $t$, i.e., $k \leq t$, the difference between our estimation and ground-truth energy reading is bounded by,
\begin{align}
\label{eqn:errbound}
    |\hat{\mathbf{e}}_{ijk} - \mathbf{e}_{ijk}^*| \leq& \alpha_i^t||\hat{\mathbf{a}}_j^t \circ \hat{\mathbf{s}}_k^t||_{(\mathbf{A}_{i}^t)^{-1}} + \alpha_j^t||\hat{\mathbf{h}}_i^t \circ \hat{\mathbf{s}}_k^t||_{(\mathbf{C}_{j}^t)^{-1}} \\
    &+  4PQR(q_2 + \epsilon_2)^{t+1} + 2PQR(q_3 + \epsilon_3)^{t+1}\nonumber
\end{align}
in which $\alpha_i^t$ and $\alpha_j^t$ are the upper bound of $||\hat{\mathbf{h}}_i^t - \mathbf{h}^*_i||_{\mathbf{A}_i^t}$ and $||\hat{\mathbf{a}}_j^t - \mathbf{a}^*_j||_{\mathbf{C}_j^t}$, and they can be explicitly calculated based on Lemma~\ref{lemma}. 
The detailed derivation is provided in our supplementary material. According to Lemma~\ref{lemma}, the last two terms on the right-hand side of the inequality are upper bounded, and their upper bound is independent from the procedure of sample selection. And the first two terms are calculated based on the currently estimated parameters in \model{}.

Based on Eq \eqref{eqn:errbound}, we compare the contribution of the selection made by \model{} in reducing its uncertainty of energy breakdown with that from any other possible selections. Denote <$x_a, y_a$> as the indices of <home, appliance> pair selected by \model{} at the end of month $t$, and <$x_o, y_o$> as the indices of any other pairs. At month $t+1$, because of different new observations introduced by these two selections, e.g., energy consumed by <$x_a$,$y_a$> and <$x_o$, $y_o$>, we have two different energy breakdown estimations. Here, we use ($\hat{\mathbf{e}}_{x_a, y_a, t+1}^A$, $\hat{\mathbf{e}}_{x_o, y_o, t+1}^A$) to represent the energy prediction after observing $e_{x_a, y_a, t+1}$, and ($\hat{\mathbf{e}}_{x_a, y_a, t+1}^O$, 
$\hat{\mathbf{e}}_{x_o, y_o, t+1}^O$) to represent the prediction if the observation is $\mathbf{e}_{x_o, y_o, t+1}$. Thus, for month $t+1$, we use $E_{A} (t+1)$ to represent the prediction error based on the selection made by \model{}, and $E_{O}(t+1)$ for the prediction error resulted from any other selections,
\begin{eqnarray*}
E_{A}(t+1) = |\hat{\mathbf{e}}_{x_a, y_a, t+1}^A - \mathbf{e}_{x_a, y_a, t+1}^*| + |\hat{\mathbf{e}}_{x_o, y_o, t+1}^A - \mathbf{e}_{x_o, y_o, t+1}^*|\\
E_{O}(t+1) = |\hat{\mathbf{e}}_{x_a, y_a, t+1}^O - \mathbf{e}_{x_a, y_a, t+1}^*| + |\hat{\mathbf{e}}_{x_o, y_o, t+1}^O - \mathbf{e}_{x_o, y_o, t+1}^*|
\end{eqnarray*}
Here, we only consider the estimation error contributed by these two pairs, because the errors from other estimations are bounded by the same result as shown in Eq \eqref{eqn:errbound}.

With \model{}, at month $t$,  <$x_a$, $y_a$> is selected as it has the highest uncertainty among all the <home, appliance> pairs.
With such a condition, and the $q$-linear convergence property of alternative least square, it can be proved that, at month $t+1$, for $\delta \in (0, 1)$, with probability at least $1 - \delta$, the
upper bound of the estimation error generated with \model{}, i.e., $UB(E_A(t+1))$, and the one generated with any other selections, i.e., $UB(E_O(t+1))$, satisfy:
\begin{equation}
\label{eqn:convergence}
    UB(E_{A}(t+1)) \leq UB(E_{O}(t+1))
\end{equation}
\begin{proof}[Proof Sketch]
In the proof, we first need to derive the upper bound of the estimation errors and then obtain the relationship between them.
The key idea for deriving the upper bound is to obtain the estimation error at month $t+1$ based on the learned parameters and the selection made at month $t$. Take the first term of $E_A(t+1)$ as an example, $|\mathbf{e}_{x_a, y_a, t+1}^A - \mathbf{e}_{x_a, y_a, t+1}^*|$ can be bounded by $\alpha_{x_a}^{t+1}||\hat{\mathbf{a}}_{y_a}^{t+1} \circ \hat{\mathbf{s}}_{t+1}^{t+1}||_{(\mathbf{A}_{x_a}^{t+1})^{-1}} + \alpha_{y_a}^{t+1}||\hat{\mathbf{h}}_{x_a}^{t+1} \circ \hat{\mathbf{s}}_{t+1}^{t+1}||_{(\mathbf{C}_{y_o}^{t+1})^{-1}} + C_1$, where $C_1$ is a constant and independent from any selections. First, we assume that the season factors change smoothly between adjacent months, e.g., $||\hat{\mathbf{s}}_k^t - \hat{\mathbf{s}}_{k+1}^t||_2 \leq ||u||_2 = \gamma$. With such an assumption and the $q$-linear convergence property, we relax the upper bound with the latent factors learned at month $t$, e.g., $\alpha_{x_a}^{t+1}||\hat{\mathbf{a}}_{y_a}^{t+1} \circ \hat{\mathbf{s}}_{t+1}^{t+1}||_{(\mathbf{A}_{x_a}^{t+1})^{-1}} 
\leq \alpha_{x_a}^{t+1}||\hat{\mathbf{a}}_{y_a}^{t} \circ \hat{\mathbf{s}}_{t}^{t}||_{(\mathbf{A}_{x_a}^{t+1})^{-1}} + C_2$, where $C_2$ is also a constant. Since $\mathbf{A}_{x_a}^{t}$ is a positive definite matrix, according to Sherman-Morrison equation, the first part of this new upper bound can be rewritten as:
\begin{equation*}
    \alpha_{x_a}^{t+1}||\hat{\mathbf{a}}_{y_a}^{t} \circ \hat{\mathbf{s}}_{t}^{t}||_{(\mathbf{A}_{x_a}^{t+1})^{-1}}^2 = \frac{\alpha_{x_a}^{t+1}||\hat{\mathbf{a}}_{y_a}^t \circ \hat{\mathbf{s}}_{t}^t||^2_{(\mathbf{A}_{x_a}^t)^{-1}}}{1 + ||\hat{\mathbf{a}}_{y_a}^t \circ \hat{\mathbf{s}}_{t}^t||^2_{(\mathbf{A}_{x_a}^t)^{-1}}},
\end{equation*}
where we use $A_{x_a}^t$ to replace $A_{x_a}^{t+1}$. The same technique can be applied to the other parts, and thus, it is easy to derive the upper bound of the estimation error:
\begin{align*}
    &E_{A}(t+1) \leq \frac{\alpha_{1}M}{\sqrt{1 + M^2}} + \frac{\alpha_2N}{\sqrt{1 +N^2}} + \alpha_1G + \alpha_2H + C \\
    &E_{O}(t+1) \leq \frac{\alpha_{1}G}{\sqrt{1 + G^2}} + \frac{\alpha_2H}{\sqrt{1 +H^2}} + \alpha_1M + \alpha_2N + C \\
\text{where}~~~
      & M =||\hat{\mathbf{a}}_{y_a}^t \circ \hat{\mathbf{s}}_{t}^t||_{(\mathbf{A}_{x_a}^t)^{-1}}{,}~~N = ||\hat{\mathbf{h}}_{x_a}^t \circ \hat{\mathbf{s}}_{t}^t||_{(\mathbf{C}_{y_a}^t)^{-1}}\\ 
      &  G =||\hat{\mathbf{a}}_{x_o}^t \circ \hat{\mathbf{s}}_{t}^t||_{(\mathbf{A}_{x_o}^t)^{-1}} {,}~~H = ||\hat{\mathbf{h}}_{y_o}^t \circ \hat{\mathbf{s}}_{t}^t||_{(\mathbf{C}_{y_o}^t)^{-1}}
\end{align*}
Therefore, according to the selection strategy defined in \model{}:
\begin{equation*}
    \alpha_1 M + \alpha_2 N \geq \alpha_1 G + \alpha_2 H, \text{ with } M \geq G \text{ and } N \geq H,
\end{equation*}
we can prove that with high probability, the upper bound of estimation error of \model{} is smaller than the one generated by any other selections.
\end{proof}

As $E_{O}(t+1)$ represents the predicted error generated by any other selections, this conclusion means our method can achieve the best prediction error reduction among all possible selections, or at least no worse than any one of them. 

\section{Empirical Evaluation}
\label{sec:exp}
We use the Dataport dataset~\cite{parson2015dataport} for evaluation. It is the largest public residential home energy dataset, which contains the appliance-level and household aggregate energy consumption sampled every minute from 2012 to 2018. While Dataport contains energy data from various cities in the U.S., we only focus on the data collected from Austin as it contains the largest set of homes (i.e., 534 homes) from a single region. 
We filter out the appliances with poor data quality (large proportion of missing values). We get 4 different datasets for each year between 2014 to 2017 containing 53, 93, 73, and 44 homes respectively and six appliances: air-conditioning (HVAC), fridge, washing machine, furnace, microwave and dishwasher. On this selected data set, we reconstruct the aggregate reading by the sum of the selected appliances \cite{kolter_2010, fhmm}.

\begin{figure}[t]
    \centering
    \includegraphics[width=1\linewidth]{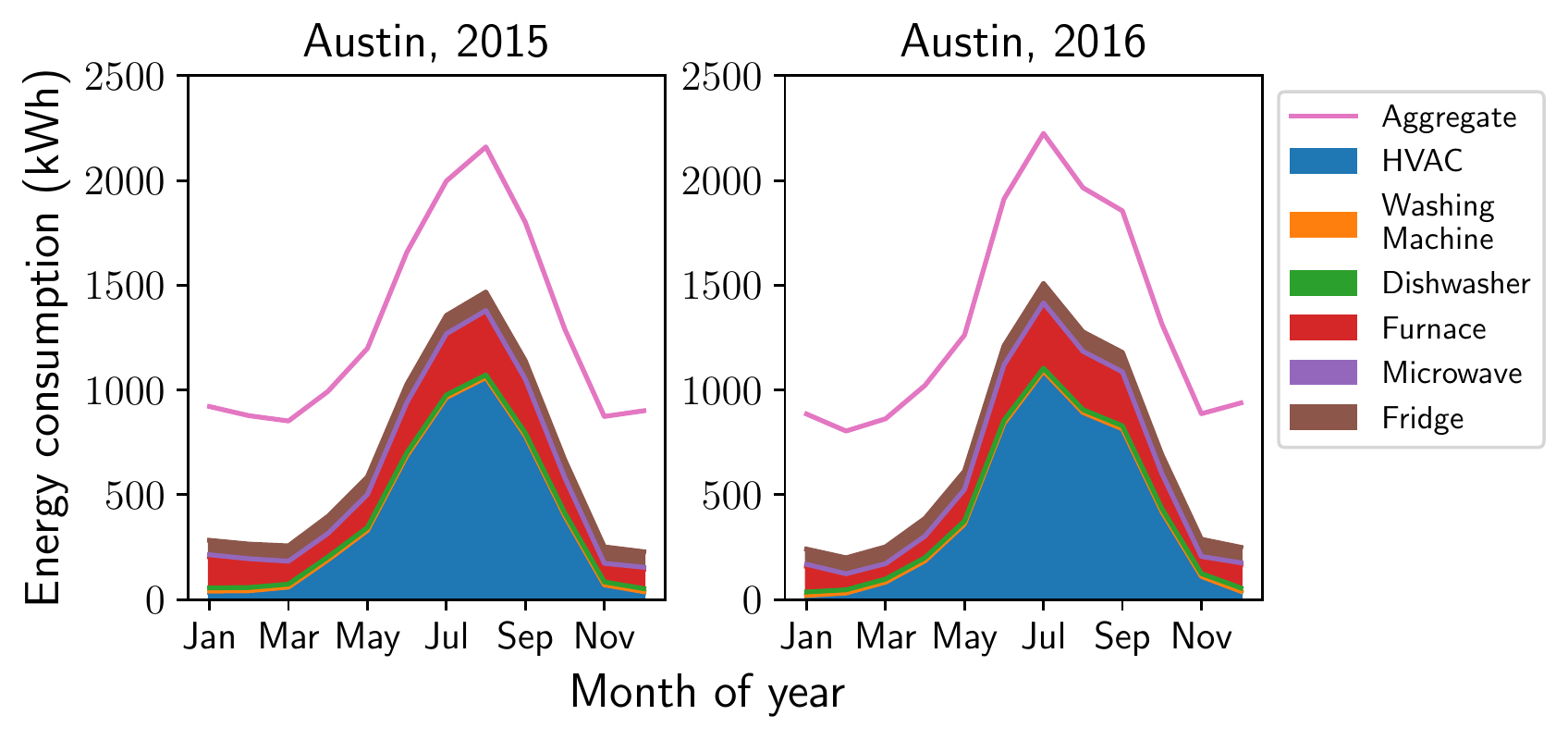}
    \vspace{-5mm}
    \caption{Energy breakdown for Austin in 2015 and 2016.}
    \vspace{-4mm}
    \label{fig:energy}
\end{figure}

An important reason for choosing these appliances is that they represent a wide variety of household energy consumption patterns. For example, season dependent (HVAC) v.s., season independent (dishwasher), background (fridge) v.s., interactive (microwave), etc. Figure~\ref{fig:energy} shows the appliance level energy consumption pattern in each month in Austin across two adjacent years. The usage patterns across months are quite similar. For example, higher consumption during summer (dominated by HVAC) and lower in winter. The energy consumption for the remaining two years while not shown is fairly similar to these results as well. This observation supports our previous assumption that the season pattern is similar across years within one geo-region, and thus, we could re-use the season patterns learned from historical information to guide future projection.

\noindent{\bf Reproducibility}
Our entire codebase, baselines, analysis and experiments can be found on
Github~\footnote{https://github.com/yilingjia/ActSense}.

\subsection{Baseline Algorithms}
In our evaluation, we use the following baselines.

\noindent{\bf $\bullet~$Random:} performs CP decomposition with ALS and selects $L$ <home, appliance> pairs uniformly random from the candidate pool. Random sampling has often been used as a baseline in prior active learning literature and also represents a feasible mechanism via which utility companies actually deploy sensors in practice.

\noindent{\bf $\bullet~$QBC:} performs CP decomposition with ALS and select pairs with Query-by-Committee (QBC) strategy. The QBC framework quantifies the prediction uncertainty based on the level of disagreement among a set of trained models. Similar to previous work~\cite{chakraborty2013active}, we perform tensor factorization using ALS with different settings of the rank parameter to form the committee. The estimation uncertainty on each <home, appliance> pair is computed by the variance across the estimates of the committee members.

\noindent{\bf $\bullet~$VBV:} performs CP decomposition with variational Bayesian inference and selects pairs according to the variance of each estimation. Variational Bayesian (VB) \cite{beal2003variational} is one of the most popularly used methods for uncertainty assessment. Prior work \cite{sutherland2013active} proposed VB for active completion in a matrix setting. In this paper, we extend their setting to tensor completion according to the work by Zhao et. al~\cite{zhao2016bayesian}. At the end of each month, VB estimates the posterior distributions of latent factors and selects the <home, appliance> pairs with the highest posterior variance in the energy estimation.

We should note that while we discussed a few other active matrix completion methods in~Section \ref{sec:relwork}, we did not use them as baselines due to their poor scalability in the energy breakdown scenario. For example, the graphical lasso algorithm~\cite{chakraborty2013active} for matrix completion needed at least 50\% of the total observations to estimate the model, and assumed the dependency between the missing values and observations follows a Gaussian distribution. A prior VB based matrix completion work~\cite{silva2012active} proposed to use the mutual information as the selection criteria. However, they model the mutual information between rows and between columns in the matrix, and select entire row or column separately. This would require us to instrument one appliance in all homes, which is clearly infeasible in practice.

\subsection{Evaluation Metric \& Setup}
Based on prior literature \cite{batra2018transferring, batra2017matrix}, we evaluate the performance of energy breakdown with root mean square error (RMSE) between the predicted appliance energy and the ground-truth on the test set. In month $t$, the ground-truth and estimated energy for appliance $j$ at home $i$ are denoted as $\mathbf{e}_{ijt}$ and $\hat{\mathbf{e}}_{ijt}$. The RMSE of appliance $j$ at month $t$ is given as,
\small
\begin{equation*}
RMSE (j, t) = \sqrt{\frac{1}{M}\sum_{i}(\mathbf{e}_{ijt} - \hat{\mathbf{e}}_{ijt})^2}
\end{equation*}
\normalsize
where $M$ represents the total number of homes in the test set. For month $t$, we use the Mean RMSE across appliances to measure a model's prediction accuracy:
\begin{equation*}
    {Mean~} RMSE(t) = \frac{\sum_j RMSE(j, t)}{N}
\end{equation*}   
where $N$ represents the number of appliances. Lower Mean RMSE indicates better energy breakdown performance.

We use 5-fold cross-validation across homes in all our experiments. The last 20\% of the train set in each fold is reserved for validation purpose. 
For each baseline and our method, the optimal parameters (e.g., rank of the tensor) are learned via an exhaustive grid search on the validation set. We first fix $\alpha_i^t$ and $\alpha_j^t$ to 0.1.
For tensor decomposition, we vary the rank of the latent dimensions from 1 to 4. With limited observations, the model could overfit, especially in the first several rounds of optimization. Thus, a larger regularization coefficient should be applied to avoid it. We choose the $\lambda$ values from \{5000, 8000, 10000\}. For the kernel function used in our uncertainty integration, we vary the window size $\sigma$ from \{1, 3, 6, 12\}.

\subsection{Experiment Results}

\subsubsection{Quality of Energy Breakdown}
\begin{figure}[t]
  \centering
  \begin{subfigure}[b]{0.5\textwidth}
    \centering
    \includegraphics[width=0.9\linewidth]{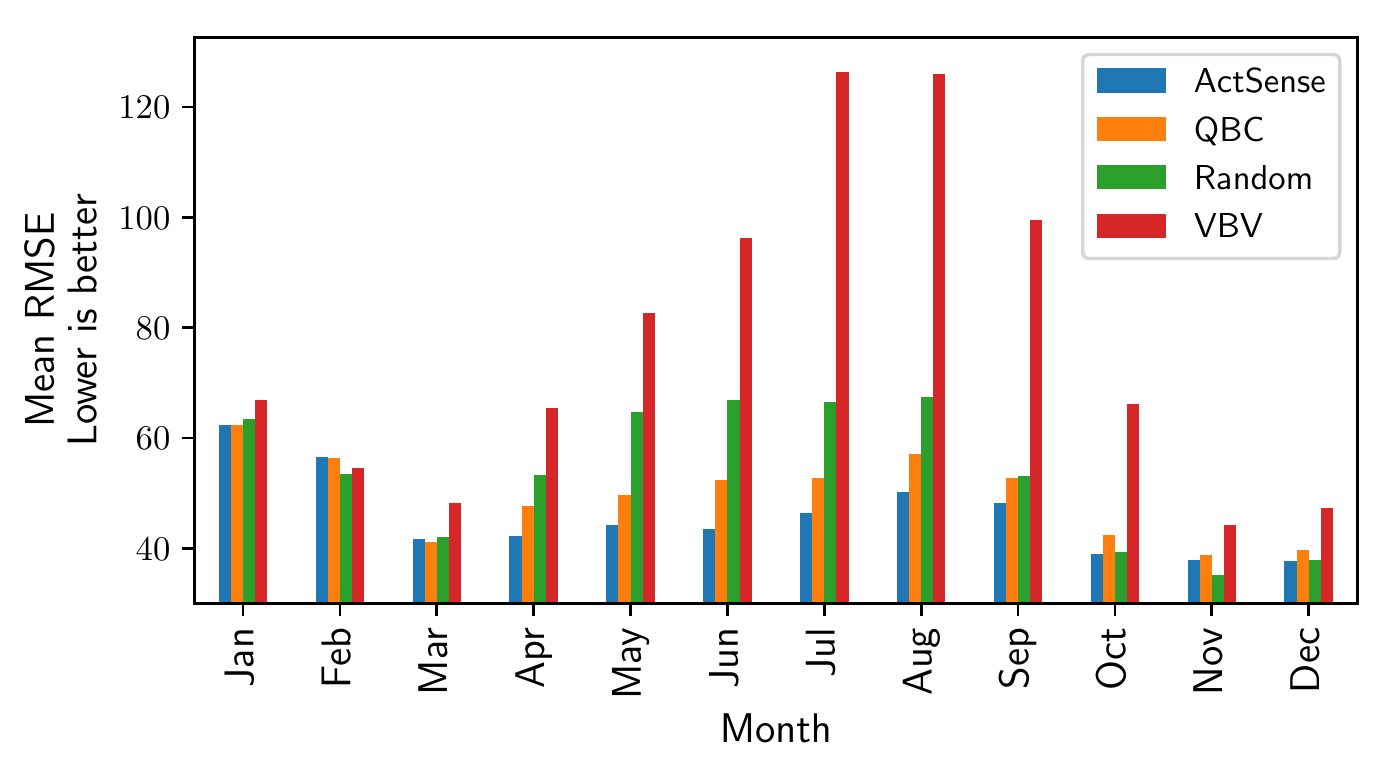}
    \vspace{-3mm}
    \caption{Mean RMSE performance across months.}
    \label{fig:montherror}
  \end{subfigure}
  \begin{subfigure}[b]{0.5\textwidth}
    \centering
    \includegraphics[width=0.9\linewidth]{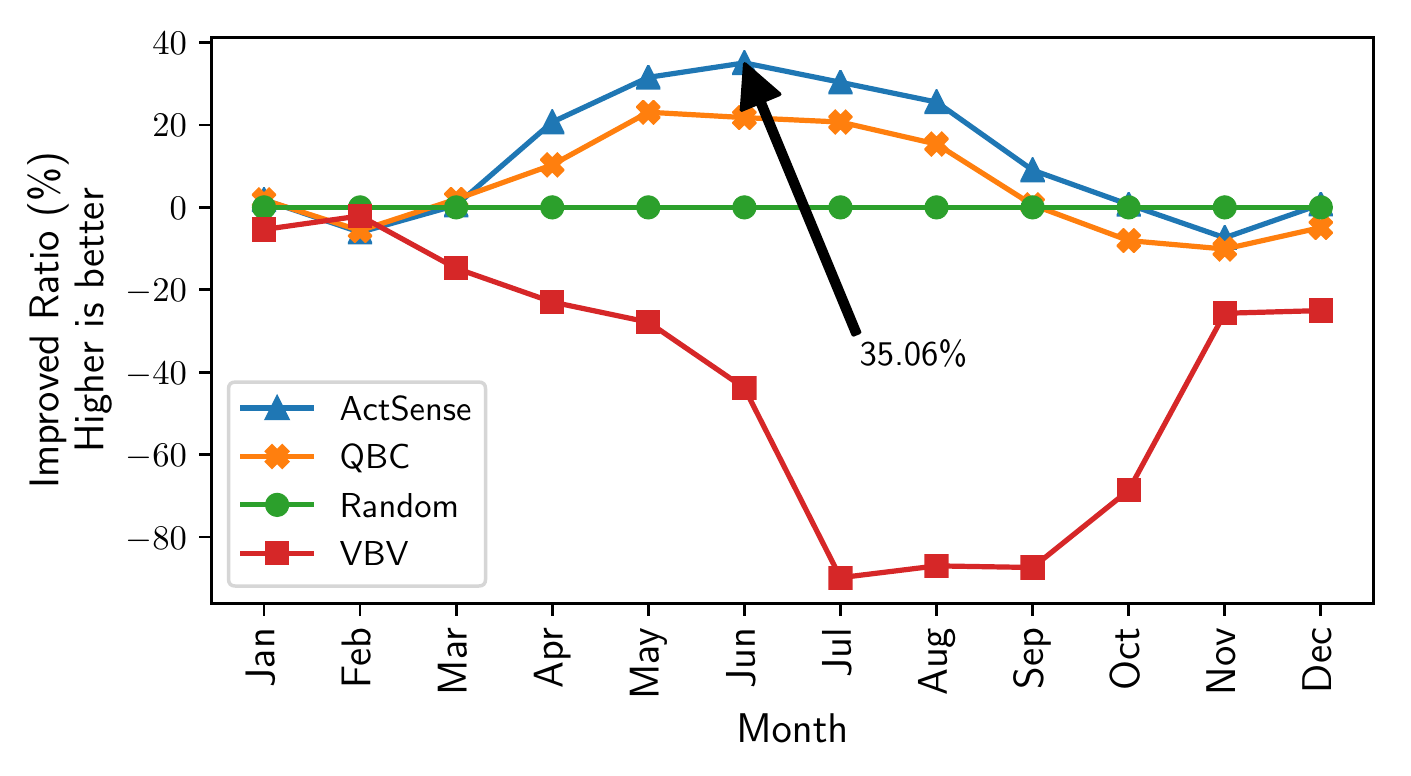}
    \vspace{-3mm}
    \caption{Relative improvement compared with random method.}
    \label{fig:improved}
  \end{subfigure}
  \vspace{-6mm}
  \caption{Our approach~\model{} gives the best energy breakdown performance across all months for the year 2015.}
  \label{fig:result}
   \vspace{-2mm}
\end{figure}

In this experiment, we compare the performance of \model{} and the baselines in monthly error, which represents the instant energy breakdown performance. 
We fixed the number of selections for each month as $L = 5$, so that for at the end of each year of our evaluation, we will have 10.75\% to 22.73\% <home,appliance> pairs instrumented across the selected four years. Figure~\ref{fig:result} shows the quality of energy breakdown in Austin, 2015. Mean RMSE across appliances for each month is reported in Figure~\ref{fig:result} (a). We can observe that our uncertainty based selection performs favourably compared to the baselines.
We also plot the relative improvement compared to the random baseline in each month in Figure~\ref{fig:result} (b). We can observe that our approach achieves the highest improvement compared to other baselines, especially in June, where the improvement is up to 35.06\%. To put this improvement into context, June and other summer months typically have the highest energy usage, where the scope and potential benefits of energy breakdown are also high.

\begin{figure}[t]
    \centering
    \includegraphics[width=0.9\linewidth]{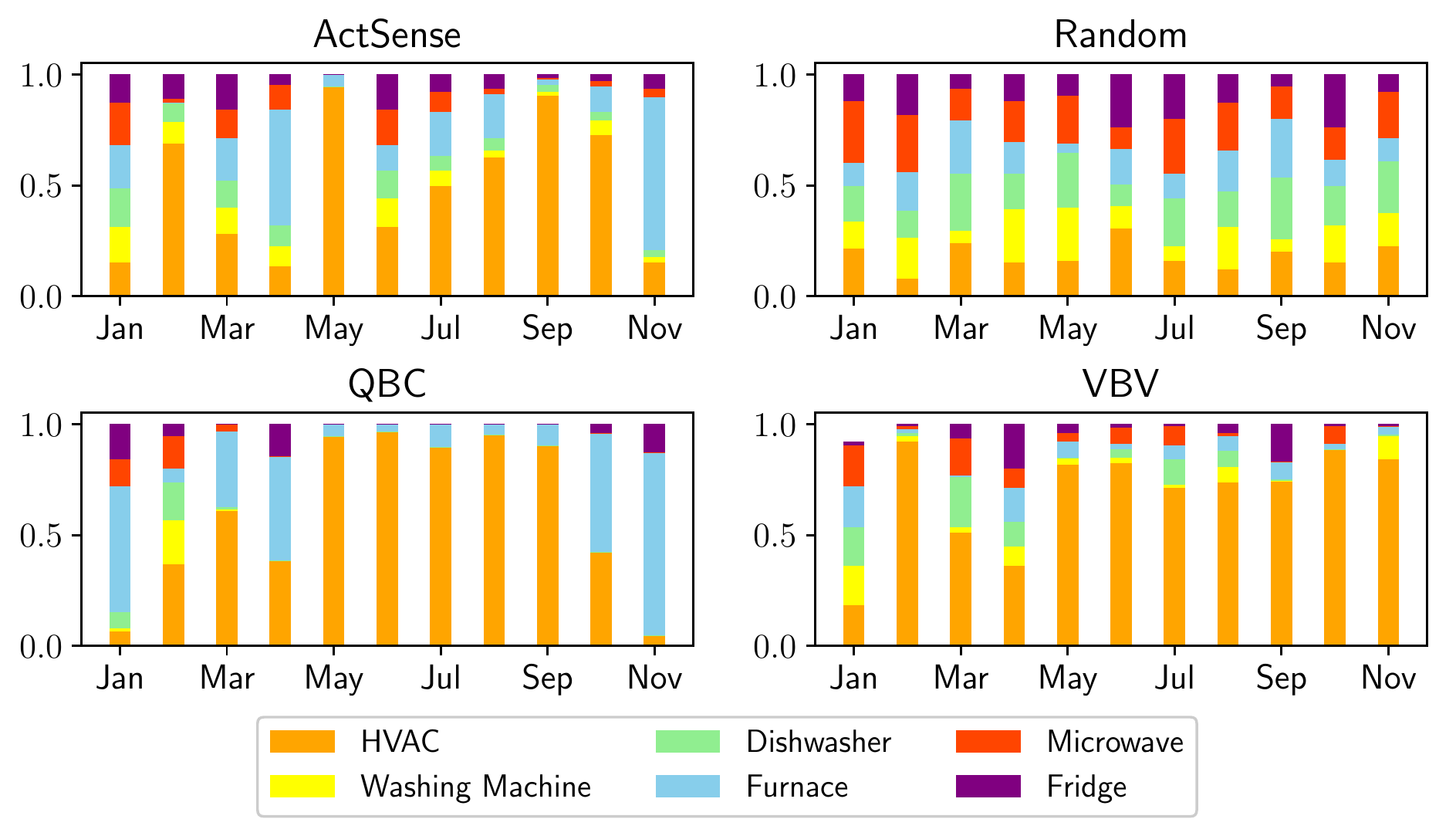}
    \caption{Selection ratio of appliances, Austin, 2015.}
    \label{fig:selected_app}
    \vspace{-6mm}
\end{figure}

We now discuss why some baselines failed to provide an accurate energy breakdown. Random selection uniformly selects the <home, appliance> pairs, ignoring  the difference in their informativeness. For example, microwave is a season independent appliance, whose usage pattern should be simpler than season dependent appliances, such as HVAC. As random selection treats all the appliances equally, additional selections made on those well-learned appliances are wasted, while the latent factors for appliances with complicated usage patterns are not well modeled. Figure~\ref{fig:selected_app} shows the selection ratio of appliances across different methods. All of them except Random selected more on HVAC or furnace, which are more difficult to model as they are highly season dependent. 
While VBV imposes the same tensor structure, it uses a different parameter estimation procedure compared to the other methods. VBV gives worse energy breakdown quality than other approaches due to its poor performance in parameter estimation. VBV estimates the distribution of each latent factor, including both mean and variance. Its model complexity is higher, and thus it can easily overfit. From its selections, we can find that it concentrated on HVAC and seldom selected other appliances. Though HVAC needs more observations to get a good estimation, VBV's selection overfits to HVAC and give worse energy performance for other appliances.
Among the baselines, QBC gives a reasonable energy breakdown quality. However, as discussed before, QBC will select the pairs which have the highest variance among the committee members. This selection strategy would perform well when data has a similar scale. But in energy breakdown, the energy readings across appliances and homes are heavily imbalanced (shown in Figure~\ref{fig:energy}). The selection is easily dominated by the appliances with high energy consumption, such as HVAC and furnace as they definitely have a higher variance than the ``minor'' appliances which in general consume lower energy, in models' predictions.
Our uncertainty based active selection strategy overcomes the problems discussed before. First, it balances the selections among the appliances. Though most of the selections are for HVAC and furnace, the other ``minor'' appliances are not ignored. Every appliance has a chance to be selected and modeled. Second, as we incorporate the temporal information into our selection strategy, \model{} can foresee the change in future and make up for the mistakes in the past. 
Intuitively, we should have more observations from HVAC in summer to accommodate its changing usage pattern. And the home and appliance pairs that the current model is still uncertain about based on historical observation will be selected with high probability.
Figure~\ref{fig:selected_app} shows that our method selects most observations of HVAC in May, while QBC selects most HVAC starting from June to September. This indicates our model foresees the future change and prepares for the sensor installation in advance. Therefore, its performance in summer is much better than the other baselines. Moreover, our approach also integrates the uncertainty from the previous months to make up for the mistakes. We can see that after May, \model{} tends to select more on the other appliances so that we can have a good model for HVAC without sacrificing the performance of others.

\begin{table}[t]
\caption{Relative improvement comparing to Random.}
\vspace{-2mm}
\label{tab:ratio}
\resizebox{0.95\linewidth}{!}{
\begin{tabular}{*{7}{|c}|}
\hline
& \multicolumn{3}{c|}{Maximum Improvement (\%)} & \multicolumn{3}{c|}{Average Improvement (\%)} \\\cline{2-7} 
& QBC & ActSense & VBV & QBC & ActSense &VBV \\ \hline
2014                  & 12.30   & 29.71 & -2.21 & 1.98  & 7.07  & -76.85 \\ \hline
2015                  & 23.09  & 35.06  & -2.07  & 5.58  & 11.88  & -41.71  \\ \hline
2016                  & 22.94  & 29.84  & -14.01  & 9.38  & 14.66  & -62.97 \\ \hline
2017                  & 7.42  & 28.76  & 0.32   & 4.48   & 12.07   & -40.82  \\\hline
\end{tabular}}
\vspace{-4mm}
\end{table}

Due to the space limit, we only report the results in other years with the best and average relative improvement across months in Table~\ref{tab:ratio}; and the detailed results are provided in our supplementary material. We can observe \model{} gives encouraging improvement over the baselines in general across these four years' evaluation.

\begin{figure}[t]
    \centering
    \includegraphics[width=0.9\linewidth]{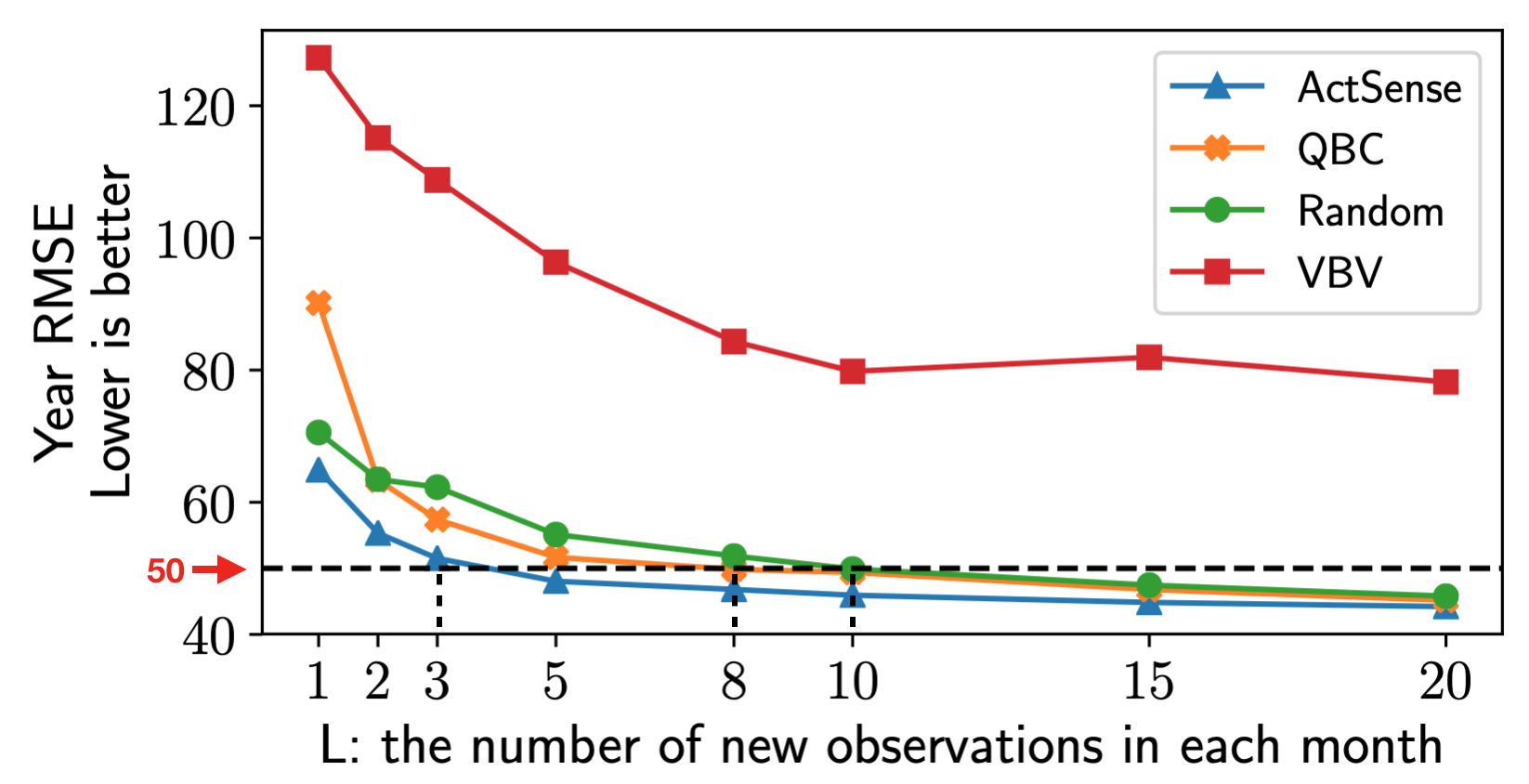}
    \caption{Performance v.s., the number of selections in each month, Austin, year 2015.}
    \label{fig:varyingk}
\end{figure}

\subsubsection{Budget size.}
As discussed before, sensor installation is expensive, and the goal of active sensor deployment is to maximize the energy breakdown accuracy while minimizing the cost of instrumentation. In this experiment, we want to explore how the budget size affects the energy breakdown performance in each method. 
We fix the hyper-parameters with the best set found in the grid search, and only vary the number of <home, appliance> pairs selected in each month from 1 to 20. With such a setting, the total number of sensors installed during one year will vary from 11 to 220. With total 93 homes and 6 appliances in the dataset from Austin in the year of 2015, this setting makes the installation ratio vary from 1.97\% to 39.43\%. We evaluate each method by its average energy breakdown performance in one year by,
\begin{equation*}
    \text{Year RMSE} = \frac{\sum^{12}_{t=1} Mean~RMSE(t)}{12}
\end{equation*}
Figure~\ref{fig:varyingk} shows our approach gives consistently better performance and faster convergence with an increasing size of the sensor installation budget. For a more practical comparison, with a fixed target of energy breakdown, say setting Year RMSE to 50, shown in the figure, our approach only needs 3 new observations per month, while QBC needs 8 and Random needs 10 new observations. This clearly demonstrates our solution's advantage in minimizing the cost of sensor deployment for improving energy breakdown quality.

\subsubsection{Temporal information}
In this section, we analyze the contribution of the temporal information incorporated into our selection strategy. In the experiment, we compare our proposed method with different uncertainty estimations: 1) Current: selects pairs based on the uncertainty on current month only; 2) Current + Future: combines the current uncertainty and the future uncertainty; 3) History + Current + Future: integrates these three types of uncertainty estimation as in our Algorithm~\ref{algo:algo2}.

\begin{table}[t]
\caption{Relative Improvement comparing to Random with different uncertainty estimation.}
\label{tab:temporal}
\begin{tabular}{lll}
\toprule
 Uncertainty Estimation    & Maximum & Mean  \\
\midrule
Current                    & 34.38\%   & 11.48\% \\
Current + Future           & 34.89\%  & 11.82\% \\
History + Current + Future & 35.06\%   & 11.88\% \\
\bottomrule
\end{tabular}
\vspace{-2mm}
\end{table}

\begin{figure}[t]
    \centering
    \includegraphics[width=1\linewidth]{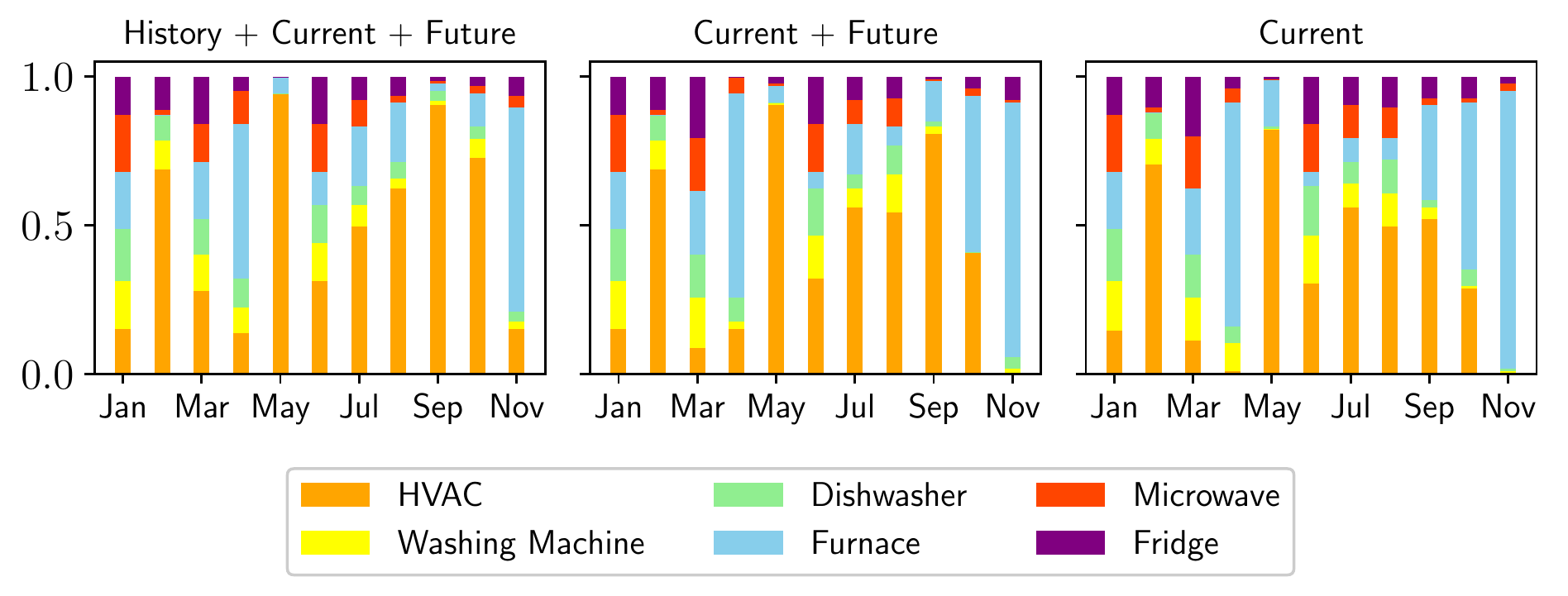}
    \caption{Selection ratio of appliances for different uncertainty measurement.}
    \label{fig:temporal}
\end{figure}

The results in Table~\ref{tab:temporal} show the best and average relative improvement of different uncertainty estimation techniques over Random across months. It can be seen that both future and historical estimated uncertainty improve the energy breakdown performance relative to Random. Furthermore, we can observe that the major contribution comes from future projection. Figure~\ref{fig:temporal} shows the selection ratio of appliances with different uncertainty estimation methods. We can notice, 1) with future projection, our approach can prepare for future changes in advance, for example, it tends to query more observations of HVAC in April and May; 2) with historical uncertainty, the model better balances between the major and minor appliances. With historical information, the algorithm can detect mistakes made earlier, and query more to make up for the same. For example, compared to the other two uncertainty measurements, the model with historical uncertainty tends to select more washing machine and HVAC at the end of  year, instead of only focusing on furnace. The results in Table~\ref{tab:temporal} indicates this knowledge retrospect helps for better energy breakdown performance.

\section{Discussions}
In addition to the promising empirical and theoretical results obtained by our proposed solution, there are a few limitations of our current work which we plan to address in the future.
\\
\noindent{\bf $\bullet$ Heterogeneous cost: } In our current setting, we assume that the cost of instrumenting different appliances are the same. However, in practice, the costs may vary due to different difficulty of instrumentation and the labor cost. The key of solving the problem is to balance the uncertainty reduction and the cost of sensor installation during the active deployment. Therefore, the selection strategy should also take the heterogeneous costs into consideration, e.g., solving a budget constrained optimization problem.
\\
\noindent{\bf $\bullet~$Dynamic installation under budget constraints: }
Our current setting assumes the number of sensors installed in each month is fixed, denoted with $L$ in the paper. However, another practical setting for sensor deployment is to consider a fixed total number of sensors. This introduces another question - how to distribute sensor deployment across months? As discussed earlier, the appliance energy consumption is highly related with season. This indicates that the complexity of usage patterns across months is different. Thus, in order to get a good estimation of energy breakdown, the number of observations required for each month should be different. Similar to the way we leveraged temporal information in our current method, we can estimate the distribution of uncertainties in future months, and dynamically change the number of sensor installation in each month according to the total budget constraint.
\\
\noindent{\bf $\bullet~$Transferable active learning: }
Our current approach only works for a single region. We assume that within one region, the structure across homes is similar due to the common design and season pattern. However, the available energy data across geo-regions is imbalanced. It is hard to generate a good energy breakdown model for a region with limited available data. We can model homogeneous or region-independent factors about the appliances, and the similarity of homes from source region and target region to actively install sensors for the target region.

\section{Conclusion}

Active collaborative sensing for sensor deployment aims to balance the performance of energy breakdown and the cost of instrumentation. The main challenge is to select the most representative or informative observations for the non-instrumented homes. To the best of our knowledge, this is the first work addressing the active sensing problem in energy breakdown. In our work, we quantify the uncertainty in the parameter estimation process for query selection. We also integrate temporal information to retrospect  history and foresee future. Our theoretical analysis and empirical evaluation prove that our approach performs favourably compared to the state-of-the-art. We believe that our method has the potential to create a paradigm shift in sensor deployments and generate positive contribution to global energy saving.

\section{Acknowledgements}
We want to thank the reviewers for their insightful comments. This work is based upon work supported by National Science Foundation under grant CNS-1646501, IIS-1553568, IIS-1718216.

\bibliographystyle{ACM-Reference-Format}
\bibliography{sample-base}

\clearpage

\section*{Supplementary}
In this supplementary document, we provide detailed proofs for Lemma~\ref{lemma} in our paper. We use the same notation as those in the paper.

\noindent{\bf Proof of Lemma ~\ref{lemma}}

By taking the gradient of the objective function defined in Eq \eqref{eqn:obj} with respect to $(\hat{\mathbf{h}}_i^t$, $\hat{\mathbf{a}}_j^t$, $\hat{\mathbf{s}}_k^t)$, and applying our model assumption that $\mathbf{e}_{ijk} = <\mathbf{h}_i, \mathbf{a}_j, \mathbf{s}_k> + \eta_{ijk}$ where $ \eta_{ijk} \sim \mathcal{N}(0,\,\sigma^{2})\,$, we have,
\begin{align*}
     \mathbf{A}^t_i(\hat{\mathbf{h}}_i^t - \mathbf{h}_i^*) 
    =& \sum_{(i, j, k) \in \Omega_t} (\hat{\mathbf{a}}_j^t \circ \hat{\mathbf{s}}_k^t)\big((\mathbf{a}_j^* \circ \mathbf{s}_k^* - \hat{\mathbf{a}}_j^t \circ \hat{\mathbf{s}}_k^t)^\top \mathbf{h}^*_i\big) \\
     &+ \sum_{(i, j, k) \in \Omega_t}\eta_{ijk}(\hat{\mathbf{a}}_j^t \circ \hat{\mathbf{s}}_k^t) - \lambda_1\mathbf{h}_i^*
\end{align*}

Therefore, we can bound the function norm of the difference between $\hat{\mathbf{h}}_i^t$ and $\mathbf{h}_i^*$ by,
\begin{align*}
        ||\hat{\mathbf{h}}_i^t - \mathbf{h}_i||_{\mathbf{A}_i^t}
        =& ||\mathbf{A}^t_i(\hat{\mathbf{h}}_i^t - \mathbf{h}_i^*)||_{(\mathbf{A}_i^t)^{-1}} \\
        =& ||\sum_{(i, j, k) \in \Omega_t} (\hat{\mathbf{a}}_j^t \circ \hat{\mathbf{s}}_k^t)((\mathbf{a}_j^* \circ \mathbf{s}_k^* - \hat{\mathbf{a}}_j^t \circ \hat{\mathbf{s}}_k^t)^\top \mathbf{h}^*_i) \\
     &+ \sum_{(i, j, k) \in \Omega_t}\eta_{ijk}(\hat{\mathbf{a}}_j^t \circ \hat{\mathbf{s}}_k^t) - \lambda_1\mathbf{h}_i^*||_{(\mathbf{A}_i^t)^{-1}} \\
     \leq& \frac{PQR}{\sqrt{\lambda_1}} \sum_{(i, j, k) \in \Omega_t} ||\mathbf{a}_j^* \circ \mathbf{s}_k^* - \hat{\mathbf{a}}_j^t \circ \hat{\mathbf{s}}_k^t||_2 \\
     &+ ||\sum_{(i, j, t)\in \Omega_t}\eta_{ijk}(\hat{\mathbf{a}}_j^t \circ \hat{\mathbf{s}}_k^t)||_{(\mathbf{A}_i^t)^{-1}} + \sqrt{\lambda_1}P
\end{align*}
where the second term on the right-hand side of the inequality can be bounded by the property of self-normalized vector-valued martingales~\cite{abbasi2011improved} as $\hat{\mathbf{h}}_i^t$, $\hat{\mathbf{a}}_j^t$, and $\hat{\mathbf{s}}_k^t$ have a finite L2 norm and $\eta_{ijk}$ has a finite variance.

For the first term, if the regularization parameter $\lambda_1$ is sufficiently large, the Hessian matrix of Eq \eqref{eqn:obj} is positive definite based on the property of alternating least square~\cite{uschmajew2012local}. The estimation of $\hat{\mathbf{h}}_i^t$, $\hat{\mathbf{a}}_j^t$, and $\hat{\mathbf{s}}_k^t$ is thus $q$-linear convergent with respect to the optimizer, which indicates that for every $\epsilon_1 > 0$, $\epsilon_2 > 0$, $\epsilon_3 > 0$, we have
\begin{align}
    ||\hat{\mathbf{h}}_i^t - \mathbf{h}_i^*||_2 &\leq (q_1 + \epsilon_1) ||\hat{\mathbf{h}}_i^{t-1} - \mathbf{h}_i^*||_2 \label{eqn:qlinear_h}\\
    ||\hat{\mathbf{a}}_j^t - \mathbf{a}_j^*||_2 &\leq (q_2 + \epsilon_2) ||\hat{\mathbf{a}}_j^{t-1} - \mathbf{a}_j^*||_2 \label{eqn:qlinear_a}\\
    ||\hat{\mathbf{s}}_k^t - \mathbf{s}_k^*||_2 &\leq (q_3 + \epsilon_3) ||\hat{\mathbf{s}}_k^{t-1} - \mathbf{s}_k^*||_2 \label{eqn:qlinear_s}
\end{align}
where $0 < q_1 < 1$, $0 < q_2 < 1$, $0 < q_3 < 1$. And for the element-wise product of two vectors, it is easy to get the conclusion that
\begin{align*}
    ||\mathbf{a}_j^* \circ \mathbf{s}_k^* - \hat{\mathbf{a}}_j^t \circ \hat{\mathbf{s}}_k^t||_2 \leq R||\hat{\mathbf{a}}_j^t - \mathbf{a}_j^*||_2 + Q ||\hat{\mathbf{s}}_k^t - \mathbf{s}_k^*||_2
\end{align*}

Therefore, we can conclude that for any $\delta > 0$, with probability at least $1 - \delta$,

\begin{equation*}
    ||\hat{\mathbf{h}}_i^t - \mathbf{h}^*_i||_{\mathbf{A}_i^t} 
    \leq \sqrt{r\ln{\frac{\lambda_1r+|\Omega_t|Q^2R^2}{\lambda_1 \cdot r \cdot \delta}}} + \sqrt{\lambda_1}P + \frac{2PQ^2R^2}{\sqrt{\lambda_1}}(G_2 + G_3)
\end{equation*}
where
\begin{align*}
\begin{split}
    G_2 = \frac{f_2(1 - f_2^{|\Omega_t|})}{1 - f_2} \,{,}\, &G_3 = \frac{f_3(1 - f_3^{|\Omega_t|})}{1 - f_3} \\
   f2 = q_2 + \epsilon_2 \,{,}\, &f3 = q_3 + \epsilon_3
\end{split}
\end{align*}
The same proof techniques apply to the proof of $||\hat{\mathbf{a}}_j^t - \mathbf{a}^*_j||_{\mathbf{C}_j^t}$ and $||\hat{\mathbf{s}}_k^t - \mathbf{s}^*_k||_{\mathbf{E}_k^t}$.

\noindent{\bf Proof of Eq~\eqref{eqn:convergence}} 

First, we give a detailed derivation of Eq~\eqref{eqn:errbound}. At month $t$, with the property of $q$-linear convergence (i.e., Eq~\eqref{eqn:qlinear_a} and Eq~\eqref{eqn:qlinear_s}), and the L2 norm constraint on the latent factors, the error between the model's estimation and ground-truth is bounded by: 
\begin{align*}
    \begin{split}
        &|\hat{\mathbf{e}}_{ijk} - \mathbf{e}_{ijk}| \\
        =& |<\hat{\mathbf{h}}_i^t, \hat{\mathbf{a}}_j^t, \hat{\mathbf{s}}_k^t> - <\mathbf{h}_i^*, \mathbf{a}_j^*, \mathbf{s}_k^*>| \\
        \leq & ||\hat{\mathbf{a}}_j^t \circ \hat{\mathbf{s}}_k^t||_{(\mathbf{A}_i^t)^{-1}}||\hat{\mathbf{h}}_i^t - \mathbf{h}_i^*||_{\mathbf{A}_i^t} + ||\hat{\mathbf{h}}_i^t \circ \hat{\mathbf{s}}_k^t||_{(\hat{\mathbf{C}}_j^t)^{-1}}||\hat{\mathbf{a}}_j^t - \mathbf{a}_j^*||_{\hat{\mathbf{C}}_j^t} \\
        &+ ||\hat{\mathbf{h}}_i^t \circ \hat{\mathbf{s}}_k^t||_2||\mathbf{a}_j^* - \hat{\mathbf{a}}_j^t||_2 + ||\hat{\mathbf{a}}_j^t \circ \hat{\mathbf{s}}_k^t - \mathbf{a}_j^* \circ \mathbf{s}_k^*||_2||\mathbf{h}_i^*||_2 \\
        \leq & ||\hat{\mathbf{a}}_j^t \circ \hat{\mathbf{s}}_k^t||_{(\mathbf{A}_i^t)^{-1}}||\hat{\mathbf{h}}_i^t - \mathbf{h}_i^*||_{\mathbf{A}_i^t} + ||\hat{\mathbf{h}}_i^t \circ \hat{\mathbf{s}}_k^t||_{(\hat{\mathbf{C}}_j^t)^{-1}}||\hat{\mathbf{a}}_j^t - \mathbf{a}_j^*||_{\hat{\mathbf{C}}_j^t} \\
        &+ 4PQR(q_2 + \epsilon_2)^{t+1} + 2PQR(q_3 + \epsilon_3)^{t+1} \\
    \end{split}
\end{align*}

With the upper bound of error, it is easy to verify that prediction error based on the selection made by ActSense is bounded by: 
\begin{align}
\label{eqn:eabound}
\begin{split}
    E_{A}(t+1) =& |\mathbf{e}_{x_a, y_a, t+1}^A - \mathbf{e}_{x_a, y_a, t+1}^*| + |\mathbf{e}_{x_o, y_o, t+1}^A - \mathbf{e}_{x_o, y_o, t+1}^*|\\
    \leq & \alpha_{x_a}^{t+1}||\hat{\mathbf{a}}_{y_a}^{t+1} \circ \hat{\mathbf{s}}_{t+1}^{t+1}||_{(\mathbf{A}_{x_a}^{t+1})^{-1}} + \alpha_{y_a}^{t+1}||\hat{\mathbf{h}}_{x_a}^{t+1} \circ \hat{\mathbf{s}}_{t+1}^{t+1}||_{(\mathbf{C}_{y_o}^{t+1})^{-1}} \\
    & + \alpha_{x_o}^{t+1}||\hat{\mathbf{a}}_{y_o}^{t+1} \circ \hat{\mathbf{s}}_{t+1}^{t+1}||_{(\mathbf{A}_{x_o}^{t+1})^{-1}} + \alpha_{y_o}^{t+1}||\hat{\mathbf{h}}_{x_o}^{t+1} \circ \hat{\mathbf{s}}_{t+1}^{t+1}||_{(\mathbf{C}_{y_o}^{t+1})^{-1}} \\ 
    & + 8PQR(q_2 + \epsilon_2)^{t+2} + 4PQR(q_3 + \epsilon_3)^{t+2}
\end{split}
\end{align}

The last two terms in the upper bound can be treated as constant, and it is the same for different selections. Thus, the major difference comes from the first four terms. 

As discussed in the paper, we assume that the season factors change smoothly between months. Thus, we assume that for two adjacent months, the difference between two corresponding factors satisfies:
\begin{equation*}
    ||\hat{\mathbf{s}}_k^t - \hat{\mathbf{s}}_{k+1}^t||_2 \leq ||u||_2 = \gamma
\end{equation*}
Thus, the first term of the upper bound can be rewritten as:
\begin{align}
\label{eqn:uncertainty_bound}
    &\alpha_{x_a}^{t+1}||\hat{\mathbf{a}}_{y_a}^{t+1} \circ \hat{\mathbf{s}}_{t+1}^{t+1}||_{(\mathbf{A}_{x_a}^{t+1})^{-1}} \nonumber \\
    =& \alpha_{x_a}^{t+1}||\hat{\mathbf{a}}_{y_a}^{t+1} \circ (\hat{\mathbf{s}}_{t}^{t+1} + u)||_{(\mathbf{A}_{x_a}^{t+1})^{-1}} \nonumber\\
    \leq & \alpha_{x_a}^{t+1}||\hat{\mathbf{a}}_{y_a}^{t+1} \circ \hat{\mathbf{s}}_{t}^{t+1}||_{(\mathbf{A}_{x_a}^{t+1})^{-1}}  + \frac{Q\gamma}{\sqrt{\lambda_1}}\nonumber\\
    \leq & \alpha_{x_a}^{t+1}||\hat{\mathbf{a}}_{y_a}^{t+1} \circ (\hat{\mathbf{s}}_{t}^{t+1} - \hat{\mathbf{s}}_{t}^{t})||_{(\mathbf{A}_{x_a}^{t+1})^{-1}}  + \alpha_{x_a}^{t+1}||\hat{\mathbf{a}}_{y_a}^{t+1} \circ \hat{\mathbf{s}}_{t}^{t}||_{(\mathbf{A}_{x_a}^{t+1})^{-1}} \frac{Q\gamma}{\sqrt{\lambda_1}}\nonumber\\
    \leq & \alpha_{x_a}^{t+1}||\hat{\mathbf{a}}_{y_a}^{t+1} \circ \hat{\mathbf{s}}_{t}^{t}||_{(\mathbf{A}_{x_a}^{t+1})^{-1}} + C_1\nonumber\\
    \leq & \alpha_{x_a}^{t+1}||(\hat{\mathbf{a}}_{y_a}^{t+1} - \hat{\mathbf{a}}_{y_a}^{t}) \circ \hat{\mathbf{s}}_{t}^{t}||_{(\mathbf{A}_{x_a}^{t+1})^{-1}} + \alpha_{x_a}^{t+1}||\hat{\mathbf{a}}_{y_a}^{t} \circ \hat{\mathbf{s}}_{t}^{t}||_{(\mathbf{A}_{x_a}^{t+1})^{-1}} +  C_1\nonumber\\
    \leq & \alpha_{x_a}^{t+1}||\hat{\mathbf{a}}_{y_a}^{t} \circ \hat{\mathbf{s}}_{t}^{t}||_{(\mathbf{A}_{x_a}^{t+1})^{-1}} + C_2
\end{align}
where $C_1 = \frac{Q\gamma}{\sqrt{\lambda_1}} + \frac{4\alpha_{x_a}^{t+1}QR(q_3 + \epsilon_3)^{t+2}}{\sqrt{\lambda_1}}$, $C_2 = C_1 + \frac{4\alpha_{x_a}^{t+1}QR(q_2 + \epsilon_2)^{t+2}}{\sqrt{\lambda_1}}$.

The first inequality is based on the property of element-wise product and the L2 norm constraint. The second and the forth inequalities are based on Cauchy-Schwarz inequality. The third and fifth inequalities hold because of the $q$-linear convergence property in Eq~\eqref{eqn:qlinear_s} and Eq~\eqref{eqn:qlinear_a}. 

With \model{}, <$x_a$, $y_a$> is selected and its reading in month $t+1$ is used to update the latent factors.
Then, according to the update procedure of the parameter defined in Section~\ref{sec:method}, and the definition of matrix norm, we can rewrite the square of the first term in Eq~\eqref{eqn:uncertainty_bound} as:
\begin{align}
    \label{eqn:u2}
    \begin{split}
        &\alpha_{x_a}^{t+1}||\hat{\mathbf{a}}_{y_a}^{t} \circ \hat{\mathbf{s}}_{t}^{t}||_{(\mathbf{A}_{x_a}^{t+1})^{-1}}^2 \\
        =& \alpha_{x_a}^{t+1}(\hat{\mathbf{a}}_{y_a}^{t} \circ \hat{\mathbf{s}}_{t}^{t})^\top(\mathbf{A}_{x_a}^{t+1})^{-1}(\hat{\mathbf{a}}_{y_a}^{t} \circ \hat{\mathbf{s}}_{t}^{t}) \\
        =& \alpha_{x_a}^{t+1}(\hat{\mathbf{a}}_{y_a}^{t} \circ \hat{\mathbf{s}}_{t}^{t})^\top(\mathbf{A}_{x_a}^{t} + (\hat{\mathbf{a}}_{y_a}^{t} \circ \hat{\mathbf{s}}_{t}^{t})(\hat{\mathbf{a}}_{y_a}^{t} \circ \hat{\mathbf{s}}_{t}^{t})^\top)^{-1}(\hat{\mathbf{a}}_{y_a}^{t} \circ \hat{\mathbf{s}}_{t}^{t})  \\
    \end{split}
\end{align}

Since $\mathbf{A}_{x_a}^{t}$ is a positive definite matrix, according to Sherman-Morrison formula, Eq~\eqref{eqn:u2} can be derived as,
\begin{equation*}
    \alpha_{x_a}^{t+1}||\hat{\mathbf{a}}_{y_a}^{t} \circ \hat{\mathbf{s}}_{t}^{t}||_{(\mathbf{A}_{x_a}^{t+1})^{-1}}^2 = \frac{\alpha_{x_a}^{t+1}||\hat{\mathbf{a}}_{y_a}^t \circ \hat{\mathbf{s}}_{t}^t||^2_{(\mathbf{A}_{x_a}^t)^{-1}}}{1 + ||\hat{\mathbf{a}}_{y_a}^t \circ \hat{\mathbf{s}}_{t}^t||^2_{(\mathbf{A}_{x_a}^t)^{-1}}}
\end{equation*}

Similar derivation can be performed for $\alpha_{y_a}^{t+1}||\hat{\mathbf{h}}_{x_a}^{t+1} \circ \hat{\mathbf{s}}_{t+1}^{t+1}||_{(\mathbf{C}_{y_a}^{t+1})^{-1}}$.

According to the update procedure of alternative least square, in Eq~\eqref{eqn:eabound}, the third and the forth term will be the same as that in month $t$. Based on Lemma~\ref{lemma}, the difference between estimated factor and the the optimal factor is upper bounded. Thus, we use $\alpha_1$, $\alpha_2$ and $\alpha_3$ to represent the upper bound of the estimation errors of each factor. Combining Eq~\eqref{eqn:eabound}, Eq~\eqref{eqn:uncertainty_bound} and Eq~\eqref{eqn:u2}, at month $t+1$, the prediction error based on selection made by \model is bounded by,
\begin{equation*}
    E_{A}(t+1) \leq \frac{\alpha_{1}M}{\sqrt{1 + M^2}} + \frac{\alpha_2N}{\sqrt{1 +N^2}} + \alpha_1G + \alpha_2H + C
\end{equation*}
where 
\begin{align*}
    \begin{split}
        M &= ||\hat{\mathbf{a}}_{y_a}^t \circ \hat{\mathbf{s}}_{t}^t||_{(\mathbf{A}_{x_a}^t)^{-1}} {,} N = ||\hat{\mathbf{h}}_{x_a}^t \circ \hat{\mathbf{s}}_{t}^t||_{(\mathbf{C}_{y_a}^t)^{-1}}\\ 
        G &= ||\hat{\mathbf{a}}_{x_o}^t \circ \hat{\mathbf{s}}_{t}^t||_{(\mathbf{A}_{x_o}^t)^{-1}} {,} H = ||\hat{\mathbf{h}}_{y_o}^t \circ \hat{\mathbf{s}}_{t}^t||_{(\mathbf{C}_{y_o}^t)^{-1}} \\
        C &= 8PQR(q_2 + \epsilon_2)^{t+2} + 4PQR(q_3 + \epsilon_3)^{t+2} \\
        &+ \frac{1}{\sqrt{\lambda_1}}(Q\gamma + 4\alpha_1QR(q_3 + \epsilon_3)^{t+2} + 4\alpha_2QR(q_2 + \epsilon_2)^{t+2}) \\
        &+  \frac{1}{\sqrt{\lambda_2}}(P\gamma + 4\alpha_1PR(q_3 + \epsilon_3)^{t+2} + 4\alpha_2PR(q_1 + \epsilon_1)^{t+2})
    \end{split}
\end{align*}

Similarly, the prediction error of any other selection can be bounded by,
\begin{equation*}
    E_{O}(t+1) \leq \frac{\alpha_{1}G}{\sqrt{1 + G^2}} + \frac{\alpha_2H}{\sqrt{1 +H^2}} + \alpha_1M + \alpha_2N + C
\end{equation*}

According to the selection strategy defined in \model{}, at month $t$, <$x_a$, $y_a$> is selected due to its highest uncertainty among all the pairs, we have :
\begin{equation*}
    \alpha_1M + \alpha_2N \geq \alpha_1G + \alpha_2H \text{, with } M \geq G \text{ and } N \geq H
\end{equation*}
Then, it is easy to get that for $\delta \in (0, 1)$, with probability at least $1 - \delta$, the upper bound of the estimation error of \model{} satisfies:
\begin{equation*}
    UB(E_{A}(t+1)) \leq UB(E_{O}(t+1))
\end{equation*}

\noindent{\bf Quality of Energy Breakdown}

Below we report the comparison between ActSense and baselines in the energy breakdown performance in datasets collected from year 2014, 2016 and 2017. 

\begin{figure}[H]
    \centering
    \includegraphics[width=0.95\linewidth]{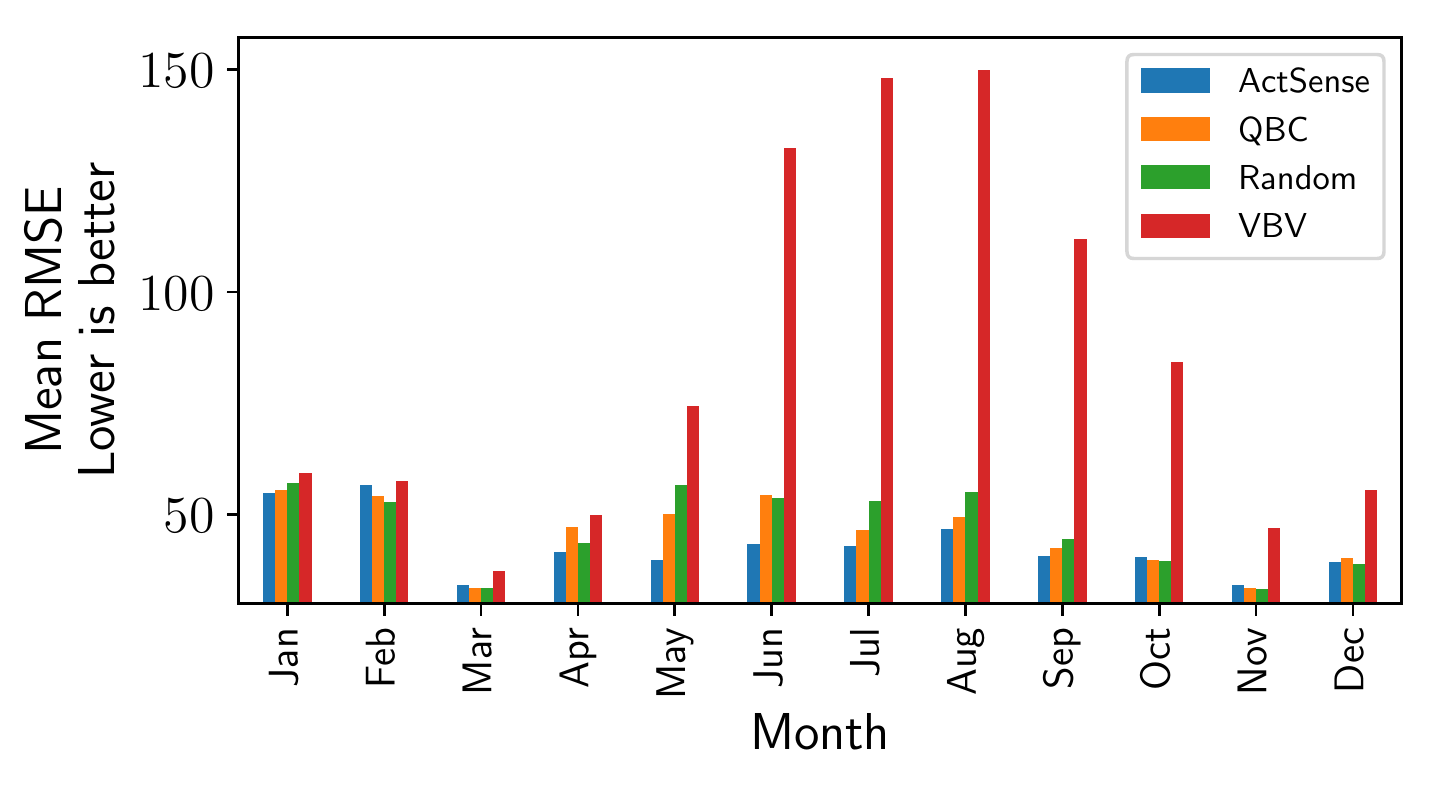}
    \caption{Mean RMSE performance, Austin, 2014.}
\end{figure}

\begin{figure}[H]
    \centering
    \includegraphics[width=0.95\linewidth]{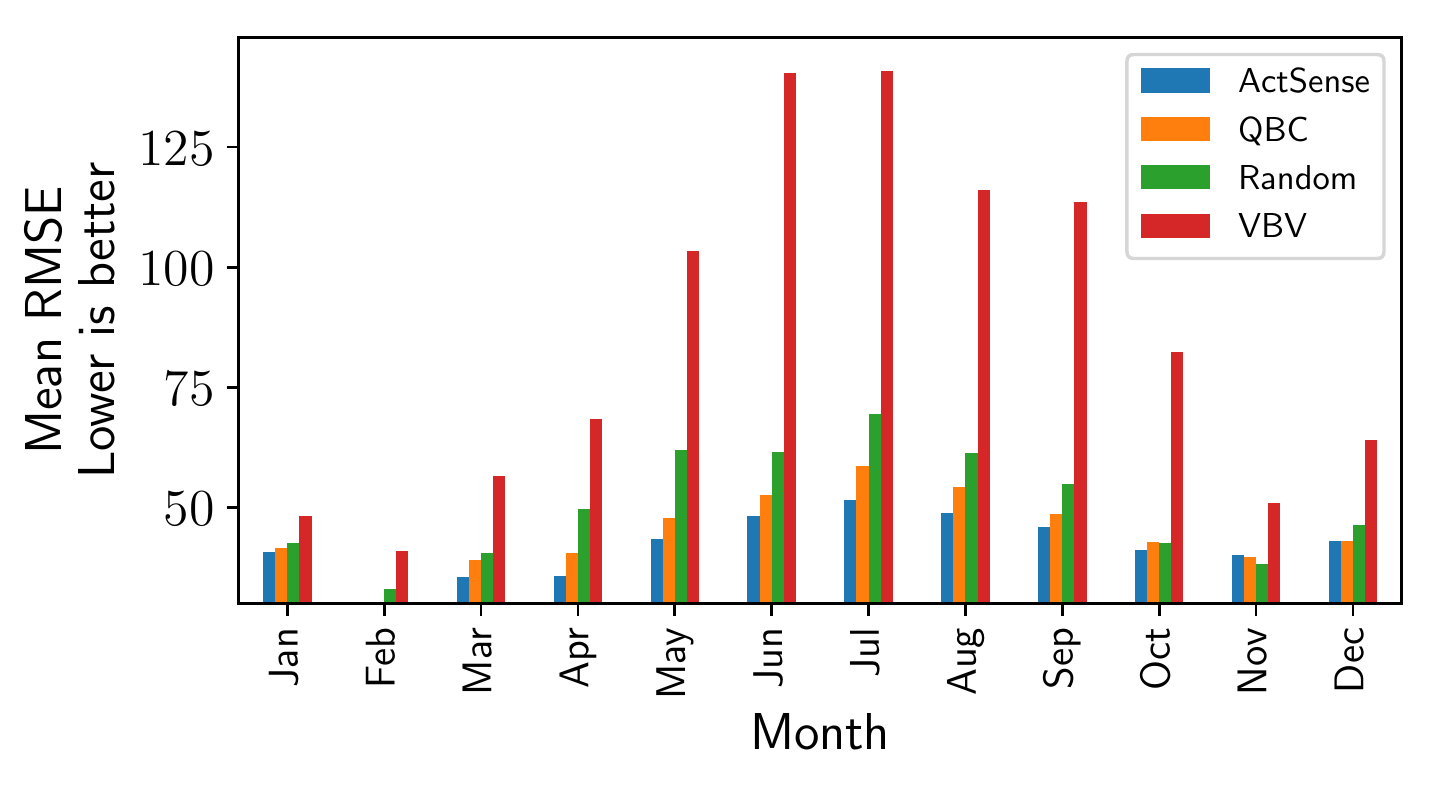}
    \caption{Mean RMSE performance, Austin, 2016.}
\end{figure}

\begin{figure}[H]
    \centering
    \includegraphics[width=0.95\linewidth]{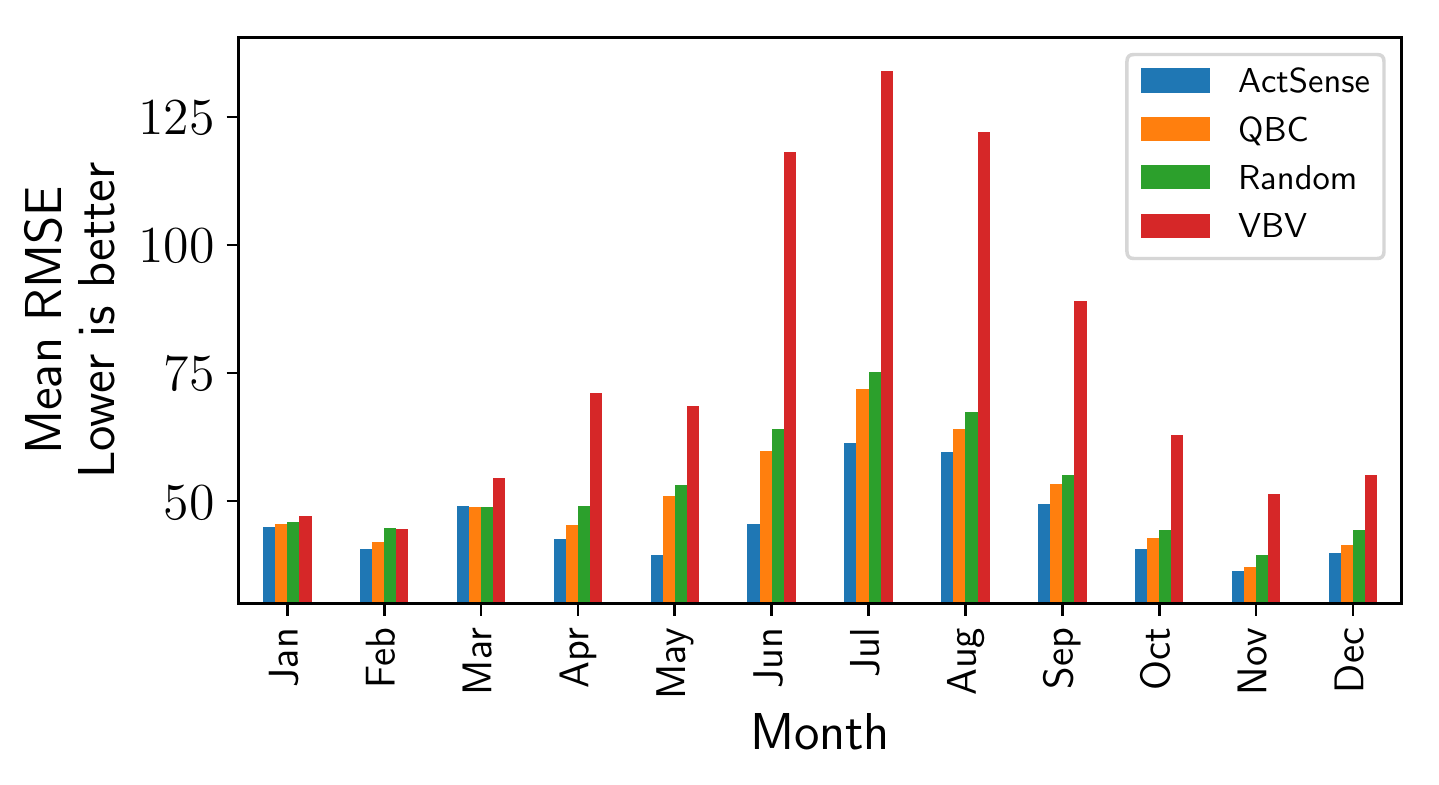}
    \caption{Mean RMSE performance, Austin, 2017.}
\end{figure}

\end{document}